\pdfoutput=1
\documentclass[11pt]{article}

\usepackage[sort&compress,square,comma,authoryear]{natbib}
\usepackage[a4paper, total={6.1in, 10in}]{geometry}

\usepackage[utf8]{inputenc}
\usepackage[T1]{fontenc}
\usepackage{url}
\usepackage{booktabs}
\usepackage{tabularx}
\newcolumntype{Y}{>{\centering\arraybackslash}X}
\newcolumntype{Z}{>{\raggedleft\arraybackslash}X}
\usepackage[parfill]{parskip}
\usepackage{nicefrac}
\usepackage{microtype}
\usepackage[table,dvipsnames]{xcolor}
\usepackage{amsmath,amssymb,amsthm}
\usepackage{bbm}                 %
\usepackage{colortbl}
\usepackage{graphicx}
\graphicspath{{./figures/}}
\usepackage{mathrsfs,mathtools}
\usepackage{caption}
\usepackage{subcaption}          %
\usepackage{wrapfig}
\usepackage{tikz}
\usetikzlibrary{calc, fadings, decorations, shapes, positioning, arrows}
\usepackage[graphicx]{realboxes}
\usetikzlibrary{arrows.meta,positioning,shapes.geometric,calc,backgrounds}

\definecolor{stageA}{HTML}{0072B2}  %
\definecolor{stageB}{HTML}{E69F00}  %
\definecolor{stageC}{HTML}{009E73}  %
\definecolor{ifpos}{HTML}{D55E00}   %
\definecolor{ifneg}{HTML}{56B4E9}   %
\definecolor{srcC}{HTML}{5A5A5A}    %

\tikzset{
  panel/.style={draw=#1, line width=0.8pt, rounded corners=4pt, fill=#1!7},
  head/.style={font=\footnotesize\bfseries, text=#1!75!black, anchor=north},
  flow/.style={-{Stealth[length=5pt,width=4pt]}, line width=0.9pt, draw=black!75},
  thinflow/.style={-{Stealth[length=4pt,width=3pt]}, line width=0.6pt, draw=black!60},
  lab/.style={font=\scriptsize, text=black!80, align=center},
  box/.style={draw=#1!80!black, fill=white, rounded corners=2pt, line width=0.6pt,
              font=\scriptsize, align=center, inner sep=3pt},
  db/.style={cylinder, shape border rotate=90, aspect=0.22, draw=black!60, fill=black!4,
             minimum width=1.05cm, minimum height=0.62cm, inner sep=1pt, font=\scriptsize},
}
\usepackage{enumitem}
\usepackage{algorithm}
\usepackage{algpseudocode}
\usepackage{makecell}
\usepackage{multirow}
\usepackage{fvextra}
\usepackage{authblk}
\DeclareMathOperator*{\argmin}{arg\,min}

\newcommand{\coderepo}{\url{https://github.com/JDIXON-2/Concept_Activation_Transport}}

\usepackage{hyperref}
\usepackage[capitalize]{cleveref}

\definecolor{salmon}{RGB}{234,153,153}
\definecolor{cornflowerblue}{RGB}{100,149,237}
\hypersetup{
    colorlinks,
    linkcolor={cornflowerblue},
    citecolor={cornflowerblue},
    urlcolor={salmon}
}

\theoremstyle{plain}

\theoremstyle{definition}

\theoremstyle{remark}

\title{Steering by Influence: Curvature Aware Data Weighting for Activation Steering}
\date{}

\author{James A.\,E. Dixon\textsuperscript{\textdagger}}
\author{Stephen J. Roberts\textsuperscript{\textdagger}}
\author{Francesco Quinzan}
\affil{Machine Learning Research Group, Department of Engineering Science\\
  University of Oxford}

\newcommand{\correspondence}{%
  {\renewcommand\thefootnote{\textdagger}%
   \footnotetext{Corresponding authors:
     \href{mailto:jamesd@robots.ox.ac.uk}{\texttt{jamesd@robots.ox.ac.uk}},
     \href{mailto:sjrob@robots.ox.ac.uk}{\texttt{sjrob@robots.ox.ac.uk}}.}}}

\begin{document}

\maketitle
\correspondence

\begin{abstract}

Inference-time steering offers cheap, fine-grained control over a language model's outputs by estimating a concept's representation in activation space and shifting activations towards it. Existing methods build these representations from activation averages over contrastive datasets. These averages incorporate unrelated concepts and noise, and are dominated by a few tokens, meaning the activation transport encodes token-level rather than thematic concepts.
In this work, we steer towards examples that most express a concept thematically, rather than towards an expectation over all. We identify these examples using influence functions, which estimate how much each data point contributes to a model's representation of a concept. Unlike simple model activation similarity, they incorporate the curvature of the model's loss landscape, allowing them to capture concept-relevant relationships beyond superficial token-level similarity.
 We then propose influence-weighted activation transport, which uses optimal transport to steer activations of non-concept text towards those of concept text, weighting concept examples by their influence scores.
We evaluate on toxicity suppression (Jigsaw), object-based concept induction (OneSec) and truthfulness induction (TruthfulQA), outperforming existing activation-transport baselines.
We track capability after steering using perplexity and MMLU accuracy, finding that our method improves steering while largely preserving model quality. We further show that influence functions capture concept-relevant information that activation-based methods miss with the two approaches ranking data points significantly differently. Together, these results demonstrate the value of curvature-aware influence information for activation steering.

\end{abstract}

\section{Introduction}

The rapid increase in size and capability of deep neural network models \cite{cottier2024rising, kaplan2020scaling} has moved the field from smaller, cheaply trained specialist models to large, expensive-to-train general models, requiring nation-state or frontier-lab resources to train \cite{EpochAIModels2026}. It has therefore become infeasible for most to train large general models from scratch, whether to adapt to a specific context, remove data, or introduce new capabilities. The field has therefore created a wide variety of post-training model adaptations, such as LoRA \cite{hu2022lora}, quantisation \cite{frantar2022gptq} and continual learning \cite{wang2024comprehensive}, to adapt models without expensive re-training. Activation steering is one such cheap post-training model adaptor that seeks to control model output by directly changing activation representations of the model at inference time, underpinned by the linear representation hypothesis \citep{park_linear_2024}. By constructing the model's activation representations of desired or undesired concepts, activation steering moves inference-time activations towards or away from these compiled representations \citep{subramani2022extracting, turner_steering_2024}. Activation steering has been achieved by using a contrastive dataset (both with and without the desired concept/behaviour) to create a contrastive gradient via differences \citep{panickssery_steering_2024, jorgensen2023meancentring}, as well as by using optimal transport maps between the two activation representation distributions as a steering map \citep{singh2024surgery, rodriguez_controlling_2024, rodriguez_LinEAS_2025}.

In another field of model interpretability, influence functions have been used to trace compromised outputs \citep{DBLP:journals/tmlr/LiLZWPS26}, modify model behaviour through training-data perturbations \citep{rosser2026infusion}, and identify thematically malign code \citep{quinzan2026detecting} based on its thematic links to other data. Using second-derivative techniques \citep{koh_understanding_2017, bae_if_2022}, thematic links between data can be identified beyond the token level \citep{grosse_studying_2023}.

Existing transport-based methods weight every sample in a class equally \citep{singh2024surgery, rodriguez_controlling_2024}, regardless of how well it encodes the concept, meaning the transport direction is noised by unrelated concepts mixed in, as well as by weak examples of the concept. Methods that do weight training data \citep{nguyen2026grains} do so at the token level rather than thematically (We define our use of the term `thematic' as: ``the underlying concept a sample expresses, independent of its specific wording''). We bridge this gap by weighting the distributions of an affine optimal transport map, in the style of \citet{rodriguez_LinEAS_2025}, using per-layer influence scores for each sample, so that the map steers towards the samples that most strongly define the concept rather than towards a noisy class mean.

\subsection{Contributions}
To the best of our knowledge, our novel contributions are categorised below. Additionally code for all is provided via our repository.\footnote{\coderepo}

\begin{itemize}
    \item  We introduce influence function weighted activation steering that uses curvature-corrected influence function scores to estimate how much each steering example contributes to steering at each model layer.
    \item We propose a contrastive, length-normalised measurement function over prompt--completion triples that constructs a contrastive estimate of a concept direction in parameter space.
    \item We use the block-diagonal structure of EK-FAC to decompose influence by layer, creating a distinct effective steering dataset at each model layer.
    \item We replace the uniform empirical measures of transport-based steering with
        influence-derived masses and exploit it in both a closed-form (IF-Act) and a jointly optimised (IF-LinEAS) steering methods. A single temperature hyperparameter interpolates between uniform and IF weighting.
\end{itemize}

\begin{figure*}[t]
  \centering
  \resizebox{\textwidth}{!}{\begin{tikzpicture}[x=1cm,y=1cm]
\draw[panel=stageA] (0,0) rectangle (4.75,4.35);
\draw[panel=stageB] (5.15,0) rectangle (10.55,4.35);
\draw[panel=stageC] (10.95,0) rectangle (16.85,4.35);
\node[head=stageA] at (2.375,-0.08) {(i) Contrastive concept direction};
\node[head=stageB] at (7.85,-0.08)  {(ii) Per-layer influence weights};
\node[head=stageC] at (13.9,-0.08)  {(iii) Influence-weighted transport};
\node[db] (Dp) at (7.08,5.2) {$\mathcal{D}^{+}$};
\node[lab, left=2pt of Dp] {target data};
\node[db] (Dm) at (12.45,5.2) {$\mathcal{D}^{-}$};
\node[lab, right=2pt of Dm] {source data};
\node[lab, anchor=north] at (2.375,4.25) {$K$ contrastive triples $(x,\,y_{+},\,y_{-})$};
\node[box=stageA, anchor=north, text width=4.1cm] (trip) at (2.375,3.9) {%
  \textit{``\dots saw an elegant tall''}\\[1pt]
  \textcolor{stageC!85!black}{$y_{+}$: \textit{striding giraffe}}\quad
  \textcolor{ifpos}{$y_{-}$: \textit{soaring cloud}}};
\draw[thinflow] (trip.south) -- (2.375,2.55);
\node[lab] at (2.375,2.35) {$\phi_i=\log p(y_{+}\!\mid x)-\log p(y_{-}\!\mid x)$};
\coordinate (o1) at (0.35,0.75);
\foreach \a/\r in {25/1.05,36/0.7,48/0.5,62/1.3} \draw[thinflow,stageA!55] (o1) -- ++(\a:\r);
\node[lab] at (0.8,0.35) {$\nabla_\theta\phi_i$};
\coordinate (o2) at (1.95,0.75);
\draw[black!35, dashed, line width=0.4pt] (o2) ++(15:0.85) arc (15:75:0.85);
\foreach \a in {25,36,48,62} \draw[thinflow,stageA!70] (o2) -- ++(\a:0.85);
\node[lab] at (2.4,0.35) {unit norm};
\coordinate (o3) at (3.5,0.75);
\draw[-{Stealth[length=6pt,width=5pt]}, line width=1.8pt, stageA] (o3) -- ++(43:0.95);
\node[font=\small, text=stageA!80!black] at (4.38,1.2) {$q$};
\node[lab] at (3.95,0.35) {mean};
\draw[flow] (1.45,1.25) -- (1.8,1.25);
\draw[flow] (3.0,1.25) -- (3.35,1.25);
\fill[ifpos!68!white] (6.000,2.030) rectangle ++(0.270,0.270);
\fill[ifneg!34!white] (6.270,2.030) rectangle ++(0.270,0.270);
\fill[ifpos!74!white] (6.540,2.030) rectangle ++(0.270,0.270);
\fill[ifneg!20!white] (6.810,2.030) rectangle ++(0.270,0.270);
\fill[ifneg!44!white] (7.080,2.030) rectangle ++(0.270,0.270);
\fill[ifneg!13!white] (7.350,2.030) rectangle ++(0.270,0.270);
\fill[ifneg!8!white] (7.620,2.030) rectangle ++(0.270,0.270);
\fill[ifneg!24!white] (7.890,2.030) rectangle ++(0.270,0.270);
\fill[ifpos!67!white] (6.000,1.760) rectangle ++(0.270,0.270);
\fill[ifpos!26!white] (6.270,1.760) rectangle ++(0.270,0.270);
\fill[ifneg!39!white] (6.540,1.760) rectangle ++(0.270,0.270);
\fill[ifneg!8!white] (6.810,1.760) rectangle ++(0.270,0.270);
\fill[ifneg!73!white] (7.080,1.760) rectangle ++(0.270,0.270);
\fill[ifpos!44!white] (7.350,1.760) rectangle ++(0.270,0.270);
\fill[ifneg!15!white] (7.620,1.760) rectangle ++(0.270,0.270);
\fill[ifneg!2!white] (7.890,1.760) rectangle ++(0.270,0.270);
\fill[ifneg!2!white] (6.000,1.490) rectangle ++(0.270,0.270);
\fill[ifneg!24!white] (6.270,1.490) rectangle ++(0.270,0.270);
\fill[ifpos!44!white] (6.540,1.490) rectangle ++(0.270,0.270);
\fill[ifneg!8!white] (6.810,1.490) rectangle ++(0.270,0.270);
\fill[ifneg!73!white] (7.080,1.490) rectangle ++(0.270,0.270);
\fill[ifpos!8!white] (7.350,1.490) rectangle ++(0.270,0.270);
\fill[ifpos!77!white] (7.620,1.490) rectangle ++(0.270,0.270);
\fill[ifneg!22!white] (7.890,1.490) rectangle ++(0.270,0.270);
\fill[ifneg!50!white] (6.000,1.220) rectangle ++(0.270,0.270);
\fill[ifpos!3!white] (6.270,1.220) rectangle ++(0.270,0.270);
\fill[ifpos!1!white] (6.540,1.220) rectangle ++(0.270,0.270);
\fill[ifneg!29!white] (6.810,1.220) rectangle ++(0.270,0.270);
\fill[ifneg!36!white] (7.080,1.220) rectangle ++(0.270,0.270);
\fill[ifpos!23!white] (7.350,1.220) rectangle ++(0.270,0.270);
\fill[ifpos!85!white] (7.620,1.220) rectangle ++(0.270,0.270);
\fill[ifneg!7!white] (7.890,1.220) rectangle ++(0.270,0.270);
\fill[ifneg!22!white] (6.000,0.950) rectangle ++(0.270,0.270);
\fill[ifpos!35!white] (6.270,0.950) rectangle ++(0.270,0.270);
\fill[ifneg!17!white] (6.540,0.950) rectangle ++(0.270,0.270);
\fill[ifpos!85!white] (6.810,0.950) rectangle ++(0.270,0.270);
\fill[ifneg!28!white] (7.080,0.950) rectangle ++(0.270,0.270);
\fill[ifneg!6!white] (7.350,0.950) rectangle ++(0.270,0.270);
\fill[ifneg!45!white] (7.620,0.950) rectangle ++(0.270,0.270);
\fill[ifneg!14!white] (7.890,0.950) rectangle ++(0.270,0.270);
\draw[white,line width=0.5pt] (6.000,0.950)--++(0,1.350);
\draw[white,line width=0.5pt] (6.270,0.950)--++(0,1.350);
\draw[white,line width=0.5pt] (6.540,0.950)--++(0,1.350);
\draw[white,line width=0.5pt] (6.810,0.950)--++(0,1.350);
\draw[white,line width=0.5pt] (7.080,0.950)--++(0,1.350);
\draw[white,line width=0.5pt] (7.350,0.950)--++(0,1.350);
\draw[white,line width=0.5pt] (7.620,0.950)--++(0,1.350);
\draw[white,line width=0.5pt] (7.890,0.950)--++(0,1.350);
\draw[white,line width=0.5pt] (8.160,0.950)--++(0,1.350);
\draw[white,line width=0.5pt] (6.000,0.950)--++(2.160,0);
\draw[white,line width=0.5pt] (6.000,1.220)--++(2.160,0);
\draw[white,line width=0.5pt] (6.000,1.490)--++(2.160,0);
\draw[white,line width=0.5pt] (6.000,1.760)--++(2.160,0);
\draw[white,line width=0.5pt] (6.000,2.030)--++(2.160,0);
\draw[white,line width=0.5pt] (6.000,2.300)--++(2.160,0);
\draw[gray!60,line width=0.4pt] (6.0,0.95) rectangle (8.160,2.300);
\node[lab, anchor=south, inner sep=1pt] (Ititle) at (7.08,2.38) {$\mathcal{I}^{(\ell)}_\phi(z_m)$};
\node[lab, rotate=90, anchor=south] at (5.98,1.625) {layer $\ell$};
\node[lab, anchor=north] at (7.08,0.92) {$z_m\in\mathcal{D}^{+}$};
\node[lab, anchor=north] at (7.08,0.5) {\textcolor{ifpos}{$\blacksquare$}\,$+$\; \textcolor{ifneg}{$\blacksquare$}\,$-$\; influence};
\draw[flow] (8.26,1.625) -- (8.72,1.625);
\node[lab, anchor=south] at (8.49,2.38) {z-score\\softmax$_\tau$};
\draw[gray!60] (8.830,2.045)--(10.190,2.045);
\fill[stageB] (8.875,2.045) rectangle ++(0.115,0.129);
\fill[stageB] (9.040,2.045) rectangle ++(0.115,0.012);
\fill[stageB] (9.205,2.045) rectangle ++(0.115,0.154);
\fill[stageB] (9.370,2.045) rectangle ++(0.115,0.012);
\fill[stageB] (9.535,2.045) rectangle ++(0.115,0.012);
\fill[stageB] (9.700,2.045) rectangle ++(0.115,0.012);
\fill[stageB] (9.865,2.045) rectangle ++(0.115,0.014);
\fill[stageB] (10.030,2.045) rectangle ++(0.115,0.012);
\draw[gray!60] (8.830,1.775)--(10.190,1.775);
\fill[stageB] (8.875,1.775) rectangle ++(0.115,0.154);
\fill[stageB] (9.040,1.775) rectangle ++(0.115,0.046);
\fill[stageB] (9.205,1.775) rectangle ++(0.115,0.012);
\fill[stageB] (9.370,1.775) rectangle ++(0.115,0.017);
\fill[stageB] (9.535,1.775) rectangle ++(0.115,0.012);
\fill[stageB] (9.700,1.775) rectangle ++(0.115,0.078);
\fill[stageB] (9.865,1.775) rectangle ++(0.115,0.014);
\fill[stageB] (10.030,1.775) rectangle ++(0.115,0.020);
\draw[gray!60] (8.830,1.505)--(10.190,1.505);
\fill[stageB] (8.875,1.505) rectangle ++(0.115,0.018);
\fill[stageB] (9.040,1.505) rectangle ++(0.115,0.012);
\fill[stageB] (9.205,1.505) rectangle ++(0.115,0.072);
\fill[stageB] (9.370,1.505) rectangle ++(0.115,0.015);
\fill[stageB] (9.535,1.505) rectangle ++(0.115,0.012);
\fill[stageB] (9.700,1.505) rectangle ++(0.115,0.025);
\fill[stageB] (9.865,1.505) rectangle ++(0.115,0.186);
\fill[stageB] (10.030,1.505) rectangle ++(0.115,0.012);
\draw[gray!60] (8.830,1.235)--(10.190,1.235);
\fill[stageB] (8.875,1.235) rectangle ++(0.115,0.012);
\fill[stageB] (9.040,1.235) rectangle ++(0.115,0.016);
\fill[stageB] (9.205,1.235) rectangle ++(0.115,0.015);
\fill[stageB] (9.370,1.235) rectangle ++(0.115,0.012);
\fill[stageB] (9.535,1.235) rectangle ++(0.115,0.012);
\fill[stageB] (9.700,1.235) rectangle ++(0.115,0.029);
\fill[stageB] (9.865,1.235) rectangle ++(0.115,0.250);
\fill[stageB] (10.030,1.235) rectangle ++(0.115,0.012);
\draw[gray!60] (8.830,0.965)--(10.190,0.965);
\fill[stageB] (8.875,0.965) rectangle ++(0.115,0.012);
\fill[stageB] (9.040,0.965) rectangle ++(0.115,0.040);
\fill[stageB] (9.205,0.965) rectangle ++(0.115,0.012);
\fill[stageB] (9.370,0.965) rectangle ++(0.115,0.250);
\fill[stageB] (9.535,0.965) rectangle ++(0.115,0.012);
\fill[stageB] (9.700,0.965) rectangle ++(0.115,0.012);
\fill[stageB] (9.865,0.965) rectangle ++(0.115,0.012);
\fill[stageB] (10.030,0.965) rectangle ++(0.115,0.012);
\node[lab, anchor=south, inner sep=1pt] at (9.51,2.38) {$w^{(\ell)}$};
\node[lab, anchor=north] at (9.51,0.92) {$\textstyle\sum_m w^{(\ell)}_m=1$};
\draw[flow] (4.5,1.45) -- (4.95,1.45) |- (Ititle.west);
\draw[flow] (Dp.south) -- (Ititle.north);
\node[lab, anchor=west] at (7.13,3.55) {$\nabla_\theta\mathcal{L}(z_m)$};
\draw[stageC!80!black,opacity=0.75,line width=0.45pt] (11.550,2.760)--(11.550,1.343);
\draw[stageC!80!black,opacity=0.75,line width=1.05pt] (11.550,2.760)--(11.910,1.433);
\draw[stageC!80!black,opacity=0.75,line width=1.50pt] (11.910,2.760)--(11.910,1.433);
\draw[stageC!80!black,opacity=0.75,line width=0.15pt] (12.270,2.760)--(11.910,1.433);
\draw[stageC!80!black,opacity=0.75,line width=0.72pt] (12.270,2.760)--(12.270,1.359);
\draw[stageC!80!black,opacity=0.75,line width=0.63pt] (12.270,2.760)--(12.630,1.443);
\draw[stageC!80!black,opacity=0.75,line width=1.50pt] (12.630,2.760)--(12.630,1.443);
\draw[stageC!80!black,opacity=0.75,line width=0.93pt] (12.990,2.760)--(12.630,1.443);
\draw[stageC!80!black,opacity=0.75,line width=0.57pt] (12.990,2.760)--(12.990,1.359);
\draw[stageC!80!black,opacity=0.75,line width=0.15pt] (13.350,2.760)--(12.990,1.359);
\draw[stageC!80!black,opacity=0.75,line width=1.35pt] (13.350,2.760)--(13.350,1.388);
\fill[srcC] (11.550,2.850) circle (0.075);
\fill[srcC] (11.910,2.850) circle (0.075);
\fill[srcC] (12.270,2.850) circle (0.075);
\fill[srcC] (12.630,2.850) circle (0.075);
\fill[srcC] (12.990,2.850) circle (0.075);
\fill[srcC] (13.350,2.850) circle (0.075);
\fill[stageB] (11.550,1.250) circle (0.063);
\fill[stageB] (11.910,1.250) circle (0.153);
\fill[stageB] (12.270,1.250) circle (0.079);
\fill[stageB] (12.630,1.250) circle (0.163);
\fill[stageB] (12.990,1.250) circle (0.079);
\fill[stageB] (13.350,1.250) circle (0.108);
\node[lab, anchor=west] at (13.45,2.85) {$a_{(j)}\!\sim\!\mu$};
\node[lab, anchor=west] at (13.45,1.25) {$b_{(k)}\!\sim\!\nu$};
\node[font=\small, text=stageC!80!black] at (11.3,2.05) {$\Gamma$};
\node[lab, anchor=north] at (12.45,0.92) {uniform source $\to$\\IF-weighted target};
\draw[flow] (Dm.south) -- (12.45,3.05);
\node[lab, anchor=west] at (12.5,3.55) {activations};
\draw[flow] (10.3,1.25) -- (11.3,1.25);
\node[box=stageC, text width=1.95cm, anchor=north] (act) at (15.75,3.85)
  {\textbf{IF-AcT}\\closed-form\\least squares};
\node[box=stageC, text width=1.95cm, anchor=south] (lin) at (15.75,0.45)
  {\textbf{IF-linEAS}\\learned via\\$\mathcal{L}_{\mathrm{W}}$};
\draw[flow] (13.95,2.05) -- (14.58,2.05);
\draw[thinflow] (14.58,2.05) |- (act.west);
\draw[thinflow] (14.58,2.05) |- (lin.west);
\node[lab] at (15.75,2.05) {affine map\\$T_{\ell,d}(a)$};
\node[box=black, text width=1.95cm, minimum height=1.4cm, line width=0.8pt]
  (out) at (18.3,2.05) {\small\textbf{Steered LLM}\\[3pt]$a\mapsto T_{\ell,d}(a)$\\strength $\lambda$};
\draw[flow] (act.east) -- ++(0.2,0) |- ([yshift=5pt]out.west);
\draw[flow] (lin.east) -- ++(0.2,0) |- ([yshift=-5pt]out.west);
\end{tikzpicture}
}
  \caption{Overview of IF-weighted activation transport.}
  \label{fig:pipeline}
\end{figure*}
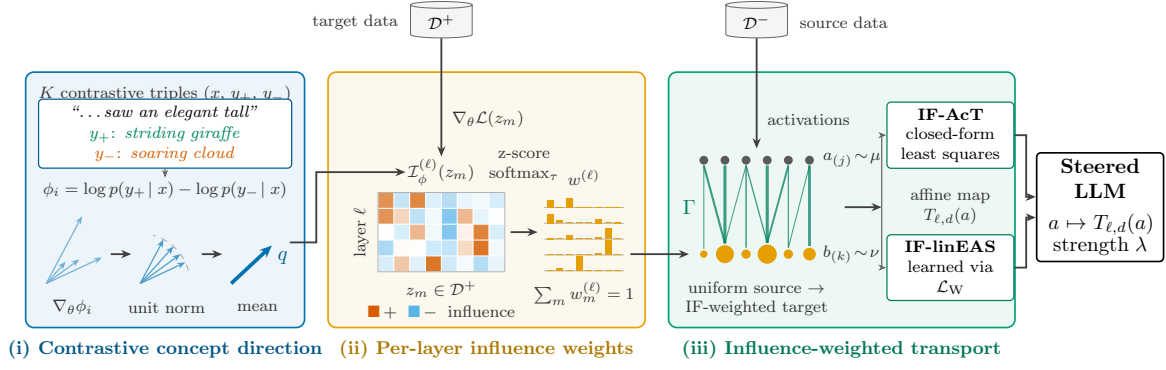

\section{Related work}

\paragraph{Activation Steering}

Activation steering was prompted by the linear representation hypothesis \citep{park_linear_2024}, which holds that high-level concepts are represented by linear directions in a model's activation space. Concepts such as refusal \citep{arditi_refusal_2024}, truthfulness \citep{burns2022discovering, marks2024geometry}, and gender \citep{belrose2023leace, bolukbasi2016man} have all been shown to be linearly represented.

Early work steered model output using expert neurons \citep{suau2022selfconditioning} and probing techniques \citep{suau2024whispering}. However, neuron-based steering is unstable due to superposition \citep{elhage2022toy}, as a single neuron may activate for multiple concepts, whereas population-level representations carry greater semantic information \citep{zou2023repe}. \citet{subramani2022extracting} and \citet{turner_steering_2024} steered using the difference between contrastive prompts, which \citet{panickssery_steering_2024} and \citet{jorgensen2023meancentring} extended to mean differences over contrastive prompt sets. \citet{wang2025act} constructed dynamic steering vectors based on the current activation's proximity to linear hallucination directions.

However, contrastive steering vectors are coarse and noisy due to their high dimensionality, with steering in a low-dimensional subspace proving more effective \citep{huang2025manifold, teo2025blessing}. They are also highly sensitive to prompt tokens, even in-distribution \citep{tan2024generalisation}, and reduce output confidence \citep{huutien2025unlearning}. All tokens in the contrastive pairs are also treated equally; \citet{nguyen2026grains} address this by using contrastive gradients to select the top-$k$ most influential tokens. An alternative line of work learns affine transformations between class-conditional distributions. \citet{singh2024surgery} derive optimal affine steering functions that minimally change representations while transporting one distribution to another. \citet{rodriguez_controlling_2024} generalise this with AcT, which fits affine transport maps grounded in optimal transport theory and forms the basis of our methodology. AcT fits each layer's map independently and sequentially; \citet{rodriguez_LinEAS_2025} instead optimise all intervention layers jointly using a Wasserstein loss, with a sparsity regulariser limiting the number and magnitude of interventions. \citet{abdullaev2026concept} relax the single-mode assumption by modelling activations as Gaussian mixtures and transporting between clusters.

\paragraph{Influence Functions}

Originating in robust statistics \citep{cook1980residuals, hampel1974influence}, influence functions (IFs) have been used to explain model predictions and identify influential training data \citep{koh_understanding_2017}. Despite concerns about their reliability in deep networks \citep{basu2020influence}, IFs approximate local training-point effects in function space \citep{bae_if_2022}. \citet{grosse_studying_2023} demonstrate that IFs capture conceptual rather than merely lexical similarity: a query about an AI being shut down was strongly influenced by training examples involving human survival despite little token overlap, with similar effects across languages and different implementations of the same algorithm. They scale this analysis to models with up to 52 billion parameters using Eigenvalue-corrected Kronecker-Factored Approximate Curvature (EK-FAC) to approximate inverse-Hessian-vector products, building on K-FAC \citep{martensgrosse2015optimizing} and subsequent extensions for weight-sharing architectures, reusable tooling, and low-precision training \citep{eschenhagen2023kronecker, osawa2023asdl, lin2024structured}.

\section{Background}

\subsection{Activation steering}

Activation steering is the practice of directly modifying a model's activations
mid-inference, either by adding to or ablating them. In our standard steering context, model $f$ is a decoder only transformer with parameters $\theta \in \mathbb{R}^{n_\theta}$. The model produces a probability distribution over the next token, $p(t_j \mid t_{<j}; \theta)$, that is sampled stochastically and is trained under cross-entropy loss $\mathcal{L}(z, \theta) = -\sum_{j=1}^{T}\log p(t_j \mid t_{<j}; \theta)$, with its training objective being the empirical average over all $N$ sequences. Formally, take the model to be a composition of transformer blocks;

Let $f$ be a decoder only transformer of $L$ blocks with parameters $\theta = (\theta_1,\dots,\theta_L)$ and residual of dimension $D$. Further, let $h_\ell(z,t) \in \mathbb{R}^D$ be the layer $\ell$ activation at token position $t$.

Here, $h_\ell(z, t) \in \mathbb{R}^D$ is the activation at token position $t$, and $\mathcal{H}$ for the set of intervention layers.
Let $\mathcal{D}^{+} = \{z_i^{+}\}_{i=1}^{N_{+}}$ be the concept dataset and $\mathcal{D}^{-} = \{z_i^{-}\}_{i=1}^{N_{-}}$ a neutral, non-concept contrastive dataset.

Each dataset induces a distribution of activations at every layer, specifically $\mu_\ell$ for $\mathcal{D}^{+}$ and $\nu_\ell$ for $\mathcal{D}^{-}$. Here steering aims to create a map that moves the activations from $\nu_\ell$ to $\mu_\ell$, while minimally disturbing activations affecting logic and broader reasoning, unrelated to the
concept. Optimal transport formalises this, creating a transport map that moves mass from $\nu_\ell$ to $\mu_\ell$  at minimal cost.

\subsection{Optimal Transport}

Following AcT \citep{rodriguez_controlling_2024} and LinEAS \citep{rodriguez_LinEAS_2025}, for a sample $z_i \in \mathcal{D}^{s}$, $s \in \{+,-\}$, with $T_i$ tokens, the activations are mean reduced over the token axis creating one mean activation for a sample for each dimension $d$ at each layer $\ell \in \mathcal{H}$,
\begin{equation*}
  \bar{h}^{s}_{\ell,d,i} = \frac{1}{T_i}\sum_{t=1}^{T_i}\big[h_\ell(z_i, t)\big]_d ,
\end{equation*}
where $[\,\cdot\,]_d$ returns the $d$-th component. Since every construction below acts on a single $(\ell, d)$ pair. For ease of notation these indices are dropped, writing $a_{(1)} \le \dots \le a_{(N_{-})}$ and $b_{(1)} \le \dots \le b_{(N_{+})}$ for the sorted source and target activations.

AcT and LinEAS constructs an affine, one-dimensional transport map for each activation dimension at specified hook points in each transformer layer. These hook points are particular locations in the decoder transformer architecture. For example, \citet{rodriguez_controlling_2024} steer the activations after the layer norms for Gemma, and the activations after the MLP up- and down-projection matrices for Llama.

\subsection{Influence Functions}

Influence functions approximate
the parameters a model would have converged to had a training point $z_m$ been upweighted by
$\epsilon$, avoiding the cost of retraining from initialisation. The effect of this perturbation on
a training set $\mathcal{D}$ is parametrised by the response function $\theta^\star(\epsilon) = \argmin_{\theta \in \mathbb{R}^{n_\theta}} \mathcal{J}(\theta, \mathcal{D}) + \epsilon \mathcal{L}(z_m, \theta)$, where $\mathcal{J}(\theta, \mathcal{D}) = \frac{1}{N}\sum_{i=1}^{N}\mathcal{L}(z_i, \theta)$. When $\epsilon > 0$ this approximates the effect of upweighting $z_m$, and by symmetry $\epsilon < 0$ approximates downweighting the sample. As per \citet{grosse_studying_2023}, $\epsilon$ is generally set to a small arbitrary value.

Applying the implicit function theorem evaluated at $\epsilon=0$, gives the approximate linearisation:
\begin{equation}
\label{eq:classical-if}
\mathcal{I}_{\theta^\star}(z_m) = \left. \frac{d\theta^\star}{d\epsilon} \right|_{\epsilon=0}
= -H^{-1} \nabla_\theta \mathcal{L}(z_m, \theta^\star),
\qquad H = \nabla^2_\theta \mathcal{J}(\theta^\star, \mathcal{D}),
\end{equation}

This derivation assumes that the model has converged to a unique optimum $\theta^\star$ at which
$H$ is positive definite, which holds for strongly convex objectives. Modern over-parametrised
networks, however, admit non-unique optima and are typically trained to parameters $\theta^s$
that are neither optimal nor stationary. Under these conditions $H$ is singular or indefinite,
and \eqref{eq:classical-if} is undefined. 

In practice $H$ is replaced by the generalised Gauss--Newton (GGN) matrix $G + \lambda I$, which is positive semi-definite whenever the loss is convex in the logits, which holds for our cross-entropy loss. With strictly positive damping, $G+\lambda I$ is positive definite.
\citet{bae_if_2022} show that this is the exact influence to the proximal Bregman response function (PBRF), which up-weights $z_m$ whilst keeping the parameters close to $\theta^s$ in weight space. The inverse-curvature term $(G+\lambda I)^{-1}$ down-weights parameter directions in which small changes would strongly alter the model's predictions on the reference data. Consequently, $\mathcal{I}_\phi(z_m)$ is large when up-weighting $z_m$ increases the target concept while requiring relatively little change to the model's existing behaviour. In this sense, the curvature correction favours concept-specific influence over directions that broadly perturb the model. As $\lambda \to \infty$, the score reduces, to plain gradient similarity.

Influence functions are not restricted to the parameters themselves. The influence of $z_m$ on
any differentiable measurement function $\phi$ of the model's outputs follows by the chain rule,
\begin{align}
\mathcal{I}_\phi(z_m) &= \nabla_\theta \phi(\theta^s)^\top \mathcal{I}_{\theta^s}(z_m) = -\nabla_\theta \phi(\theta^s)^\top (G + \lambda I)^{-1} \nabla_\theta \mathcal{L}(z_m, \theta^s).
\label{eq:if-phi}
\end{align}
\citet{grosse_studying_2023} define their measurement function as $\phi(\theta) = \log p(y \mid x; \theta)$, where $x$ is a concept-neutral prompt (e.g.\ $x =$ ``my favourite animals are'') and $y$ is a concept-rich completion (e.g.\ $y =$ ``giraffes''). 
In \eqref{eq:if-phi},
candidate data points $z_m$ that also encode the concept would increase $\log p(y \mid x; \theta)$, leading to a higher $\mathcal{I}_\phi$ score for $z_m$. This treats $\mathcal{I}_\phi(z_m)$ as the first order change in the $\phi(\theta)$, if the model were up weighted by $z_m$, whilst keeping predictions on the reference data and parameters close, as per the PBRF counterfactual.
 
The inverse damped Gauss--Newton matrix $(G + \lambda I)^{-1}$ is the most expensive part of the influence-function computation which is intractable to compute exactly at the scale of modern
language models. We therefore use the EKFAC approximation based on KFAC, details of which can be found in Appendix~\ref{app:KFAC}.

\section{Methodology}
\subsection{Targeting concepts $\nabla_\theta \phi(\theta)$}

To be able to steer towards a singular concept, we first require a scalar measure of that concept. We adapt the prompt--completion formulation of \citet{grosse_studying_2023} to steer towards our concept. 
A single prompt--completion pair provides a noisy concept direction: $\nabla_\theta \log p(y \mid x;\theta)$ captures not only the target concept, but also lexical and semantic features specific to $x$ and $y$ \citep{tan2024generalisation, huang2025manifold, teo2025blessing}. We therefore use multiple contrastive prompt--completion pairs to isolate the component consistently associated with the target concept.

To remove this non-concept noise, we take inspiration from contrastive gradient steering literature \citep{nguyen2026grains} as well as and find our concept direction as an expectation \citep{jorgensen2023meancentring} over gradients of contrastive prompt completion triples. We use $K$ contrastive triples $\{x^{(i)}, y_{+}^{(i)}, y_{-}^{(i)}\}_{i=1}^{K}$.
Each prompt has two completions, one containing the concept and the other without, e.g.\ $\{x^{(i)} =$ ``she looked out the window and saw an elegant tall'', $y_{+}^{(i)} =$ ``striding giraffe'', $y_{-}^{(i)} =$ ``soaring cloud''$\}$, resulting in a per-triple measurement function
\begin{gather*}
\phi_i(\theta) = \frac{1}{T_{+}^{(i)}} \log p\big(y_{+}^{(i)} \mid x^{(i)}; \theta\big)
            - \frac{1}{T_{-}^{(i)}} \log p\big(y_{-}^{(i)} \mid x^{(i)}; \theta\big),
\\
\log p(y \mid x; \theta) = \sum_{t=1}^{|y|} \log p\big(y_t \mid x, y_{<t}; \theta\big),
\end{gather*}
where $T_{\pm}^{(i)} = |y_{\pm}^{(i)}|$ is the number of tokens in the tokenised completion, normalising for losses being greater for longer strings. So that no single triple dominates purely through its gradient magnitude, each triples gradient is rescaled to unit norm before averaging, giving the concept direction
\begin{equation*}
q = \frac{1}{K} \sum_{i=1}^{K} \frac{\nabla_\theta \phi_i(\theta^s)}{\lVert \nabla_\theta \phi_i(\theta^s) \rVert_2},
\end{equation*}
which updates \eqref{eq:if-phi} with a mean centred contrastive completion gradient for the concept direction:
\begin{equation*}
\mathcal{I}_\phi(z_m) = -\, q^\top (G + \lambda I)^{-1} \nabla_\theta \mathcal{L}(z_m, \theta^s).
\end{equation*}
Visualisations of this can be found in Appendix~\ref{app:triples_umap}. To evaluate whether contrastive prompts improve discrimination of the target concept, we run a simple classification experiment via influence score, comparing the IF scores of the plain prompt--completion regime with those of our contrastive triples. Details for contrastive prompt comppleations dataset generation can be found in Appendix~\ref{synth-data}. Results can be found in Appendix~\ref{app:tox_classificaiton}.

\subsection{Influence scores}
Because $G \approx     \operatorname{blockdiag}\big(G_p\big)_{p \in \mathcal{P}}$ is block-diagonal over weight matrices, the influence score inherits this structure and split into a sum of independent per-matrix contributions for each layer. 
This allows us to attribute influence to a data point at a given layer in the network.

Following \citet{grosse_studying_2023}, we restrict our analysis to MLP parameters and group them by transformer block $\ell=1,\dots,L$, writing $\theta=(\theta_1,\dots,\theta_L)$, where $\theta_\ell=\{W_p\}_{p\in\mathcal{P}_\ell}$ and $W_p\in\mathbb{R}^{d_{\mathrm{out}}\times d_{\mathrm{in}}}$. Here, $\mathcal{P}_\ell$ indexes the up- and down-projection matrices of the $\ell$-th MLP block, with biases absorbed into $W_p$ by augmenting the input with a constant.
The influence score of a candidate point  is then a sum of the inner products per layer, and grouping those terms by the block they belong to gives a decomposition over layers,
\begin{equation*}
  \mathcal{I}_\phi(z_m) = \sum_{\ell=1}^{L} \mathcal{I}_\phi^{(\ell)}(z_m) = -\sum_{\ell=1}^{L}\sum_{p \in \mathcal{P}_\ell}
      \Big\langle (G_p + \lambda_p I)^{-1} q_p,\;
      \nabla_{W_p} \mathcal{L}(z_m, \theta^s) \Big\rangle_F
\end{equation*}
with $\langle \cdot , \cdot \rangle_F$ the Frobenius inner product.
Each term measures how strongly the loss gradient of $z_m$ at block $\ell$ aligns with the preconditioned concept direction at that same block. A point can therefore be highly influential at one depth and negligible at another, and we exploit this by weighting points per-layer, steering with a different effective dataset at each depth rather than a single global one.
We interpret this layer-wise influence as a measure of how strongly each candidate example contributes to the target concept at that layer. We therefore use it to weight the candidate's activation in the target distribution, assigning greater transport mass to activations associated with more concept-influential examples.

\subsection{Data point Influence Weighting}

Influence function weighting scores need to be $w^{(\ell)}_m \ge 0$, $\sum_m w^{(\ell)}_m = 1$. Currently, the layer-wise scores $\mathcal{I}^{(\ell)}_\phi(z_m) \in \mathbb{R}$ over the $N_{+}$ points $z_m \in \mathcal{D}^{+}$ are heavily tailed. To put the scores in a similar range and standard deviation we first z-score,
\begin{equation*}
  \hat{\mathcal{I}}^{(\ell)}_m = \frac{\mathcal{I}^{(\ell)}_\phi(z_m) - \bar{\mathcal{I}}^{(\ell)}}{\sigma^{(\ell)}},
  \qquad
  \bar{\mathcal{I}}^{(\ell)} = \frac{1}{N_{+}}\sum_{k=1}^{N_{+}} \mathcal{I}^{(\ell)}_\phi(z_k),
  \qquad
  \sigma^{(\ell)} = \sqrt{\frac{1}{N_{+}}\sum_{k=1}^{N_{+}} \big(\mathcal{I}^{(\ell)}_\phi(z_k) - \bar{\mathcal{I}}^{(\ell)}\big)^2},
\end{equation*}
which gives the scores scale invariance. This normalisation preserves relative influence between examples within each layer while removing differences in absolute scale across layers. We therefore use influence here to determine which examples to emphasise at each layer, rather than the relative steering strength between layers.
A softmax with temperature $\tau > 0$ over these scores,$w^{(\ell)}_m(\tau)$
allows us to then be able to tune this weighting distribution between the extremes of a point mass on the most influential point ($\tau \to 0$) and a uniform distribution over all ($\tau \to \infty$). Tuning $\tau$ therefore gives us robustness to IF score outliers. A study on the effective sample size can be found in Appendix~\ref{app:ESS_temperature}.

\subsection{Weighted Optimal Transport}
\label{sec:w-o-t}
Once the influence scores per data point are calculated, we need to inject them into an activation steering method that can be decomposed per layer. To keep inference cheap \citep{chewi2025statistical}, we follow the per-dimension affine maps of \citet{rodriguez_controlling_2024}, avoiding Sinkhorn or linear-programming solvers and decompose the transport map into computationally efficient univariate maps \citep{peyre_computational_2019}

For convex costs the optimal 1-D coupling is monotone \citep{peyre_computational_2019}.
therefore, where $c_k=\sum_{i\le k}w_i$ ($c_0=0$), source sample $j$ occupies
$(\frac{j-1}{N_-},\frac{j}{N_-}]$ on the quantile axis and target sample $k$
occupies $(c_{k-1},c_k]$ with the mass moved between them being their overlap,
\begin{equation}\label{eq:gamma}
\Gamma_{jk}=\big[\min(\tfrac{j}{N_-},c_k)-\max(\tfrac{j-1}{N_-},c_{k-1})\big]_+ .
\end{equation}
This needs only a sort (complexity $\mathcal{O}(n\log n)$) and reduces to pairing $a_{(j)}$ with
$b_{(j)}$ when weights are uniform and source and target sets are the same size, $N_-=N_+$.

\paragraph{Closed form}
Following AcT \citep{rodriguez_controlling_2024}, we introduce influence-weighted activation transport (IF-Act). 
Formally, at hook $h$ and activation dimension $d$, the affine transport of an activation $a$ is
\begin{equation}
  T_h^{(d)}(a) = a + \gamma\,(\alpha_h a + \beta_h - a),
  \label{eq:if_lin_transport}
\end{equation}
where $\alpha_h$ and $\beta_h$ are solved in closed form from the sorted activation distributions. The parameter $\gamma$ scales the intervention linearly, controlling the steering strength.

We construct the IF-weighted closed form as follows. The closed form is the least-squares fit under the quantile coupling $\Gamma$. Here $(\alpha_h, \beta_h) = \operatorname*{arg\,min}_{\alpha,\beta}\; \mathbb{E}_{(a,b)\sim\Gamma}\big[(\alpha a + \beta - b)^2\big]$,

which is solved by least-squares regression:
\begin{equation*}
  \alpha_h = \frac{\operatorname{Cov}_\Gamma(a,b)}{\operatorname{Var}_{\nu}(a)} = \frac{\textstyle\sum_{j,k}\Gamma_{jk}(a_j - \bar{a})(b_k - \bar{b})}{\tfrac{1}{N_-}\textstyle\sum_{j}(a_j - \bar{a})^2}, \qquad
  \beta_h = \bar{b} - \alpha_h \bar{a},
\end{equation*}
where $\bar{a} = \tfrac{1}{N_-}\textstyle\sum_{j} a_j, \qquad \bar{b} = \textstyle\sum_{k} w_k b_k.$
Note, that $w_k$ is the layer $h$ IF weighting of  the data point that produced $b_k$. As only the target distribution $B$ is weighted, we need only weight the covariance and the target activation mean. In the later experiments, we weight the source distribution when suppressing a concept, changing $\bar{a}$ to be a weighted mean and $\bar{b}$ to be a uniform mean. Further details of the unweighted steering algorithm can be found in \citet{rodriguez_controlling_2024}.

\paragraph{Open form}

We modify a second steering method, LinEAS \citep{rodriguez_LinEAS_2025}, to obtain \textsc{IF-LinEAS}. This allows us to examine whether the effect of influence weighting generalises across steering methods rather than being specific to IF-Act. IF-LinEAS uses the same influence-weighted transport plan $\Gamma$ defined above, but parameterises the transport at each hook point $\ell \in \mathcal{H}$ and dimension $d \in \{1,\dots,D\}$ by an affine map where $\alpha_{\ell,d},\beta_{\ell,d} \in \mathbb{R}$ are learnable parameters. We collect the parameters across all hook points and dimensions as $\Theta = \left\{ \alpha_{\ell,d},\, \beta_{\ell,d} \;\middle|\; \ell \in \mathcal{H},\; d \in \{1,\dots,D\} \right\}$.

All parameters in $\Theta$ are learned jointly by backpropagation. For each $(\ell,d)$, the source activation $a^{\ell,d}_{(j)}$ is mapped in the same manner as \eqref{eq:if_lin_transport}.
Using the weighted coupling $\Gamma$ between the source and target empirical measures, we define the corresponding one-dimensional Wasserstein transport objective as
\begin{equation}
  \mathcal{L}_{\mathrm{W}}(\Theta)
  \;=\;
  \sum_{\ell \in \mathcal{H}}
  \sum_{d=1}^{D}
  \sum_{j=1}^{N_{-}}
  \sum_{k=1}^{N_{+}}
  \Gamma^{\ell,d}_{jk}
  \left|
    T_{\ell,d}\!\left(a^{\ell,d}_{(j)}\right)
    -
    b^{\ell,d}_{(k)}
  \right|,
\end{equation}
and optimise $ \Theta^{*}=\operatorname*{arg\,min}_{\Theta}\mathcal{L}_{\mathrm{W}}(\Theta)$. Thus, influence weighting enters IF-LinEAS through $\Gamma^{\ell,d}$, while the affine parameters governing the intervention are optimised jointly across all dimensions and hook points. Further details on LinEAS's implementation and sparsity regulariser can be found in \citet{rodriguez_LinEAS_2025}.

\section{Experiments and Results}
We now compare our weighted methods against their non-weighted counterparts to test whether IF weighting produces a more effective form of steering. We experiment on two open-weight decoder-only models, Gemma-2-2B \citep{gemmateam2024gemma2} and Llama-3.2-3B \citep{dubey2024llama3}, hereafter Gemma and Llama. Hook points are listed in Appendix~\ref{sec:hook-points}. To compute our IF weightings we build the approximate Gauss--Newton Hessian, using an approximation of the model's training data: 100 extracts of Wikipedia articles taken from the Wikimedia dumps \cite{wikidump}.

\paragraph{General concept Induction}
\label{sec:onesec}
The OneSec dataset \citep{scarlini2019onesec} contains sentences for, object-based concepts. We use the extract of OneSec from \cite{rodriguez_controlling_2024}, covering the seven concepts \{\textit{baby, balloon, book, church, cloud, flower, football}\}, each with $\sim 700$ sentences. For consistency, we follow the methodology of \cite{rodriguez_controlling_2024} and build a separate contrastive dataset for each concept: $D_+$ consists of all $\sim 700$ sentences containing the target concept, and $D_-$ is a random sample of 700 sentences drawn from the other six concepts.
To evaluate, we generate from the steered model 
and use an LLM-as-a-judge \cite{zheng2023judging} 
to decide whether each generation refers to the target concept. The judge is given the concept's WordNet definition \cite{fellbaum1998wordnet} and answers \textit{yes} or \textit{no}. We report concept presence as the fraction of generations judged \textit{yes}, $p(\text{yes})$. Further generation and LLM Judge details are given in Appendix~\ref{app:llm_judge}.
Because these concepts are shallow and object-based, we expect concept induction here to be largely token-driven, and therefore less suited to our deeper, thematically driven steering.

\begin{table}[t]
  \caption{OneSec concept induction: average over concepts. Steering on all layers; steered values at $\lambda$ = 1, the unsteered row is $\lambda$ = 0. PPL columns: Flu. = fluency perplexity (of the generated continuation), Wiki = held-out WikiText perplexity. Each value is the mean over OneSec concepts.}
  \label{tab:onesec-average}
  \centering
  \small
  \setlength{\tabcolsep}{4pt}
  \renewcommand{\arraystretch}{0.9}
  \setlength{\aboverulesep}{0.2ex}
  \setlength{\belowrulesep}{0.2ex}
  \begin{tabular}{lcccc}
    \toprule
     & & \multicolumn{2}{c}{\textbf{PPL} $\downarrow$} & \\
    \cmidrule(lr){3-4}
    \textbf{Method} & \textbf{Concept score (LLM)} $\uparrow$ & \textbf{Flu.} & \textbf{Wiki} & \textbf{MMLU} $\uparrow$ \\
    \midrule
    \textit{Gemma-2-2B (unsteered)} & 0.018 {\scriptsize$\pm$0.019} & 6.58 {\scriptsize$\pm$0.00} & 32.30 {\scriptsize$\pm$0.00} & 0.549 {\scriptsize$\pm$0.000} \\
    \cmidrule(lr){1-5}
    Act & \underline{0.708} {\scriptsize$\pm$0.211} & 11.24 {\scriptsize$\pm$2.78} & 91.18 {\scriptsize$\pm$78.23} & 0.466 {\scriptsize$\pm$0.031} \\
    IF-Act & \textbf{0.721} {\scriptsize$\pm$0.194} & 32.27 {\scriptsize$\pm$20.74} & 1010.35 {\scriptsize$\pm$927.72} & 0.267 {\scriptsize$\pm$0.050} \\
    LinEAS & 0.544 {\scriptsize$\pm$0.243} & \textbf{8.38} {\scriptsize$\pm$1.08} & \textbf{55.36} {\scriptsize$\pm$21.98} & \textbf{0.499} {\scriptsize$\pm$0.024} \\
    IF-LinEAS & 0.584 {\scriptsize$\pm$0.274} & \underline{8.49} {\scriptsize$\pm$1.00} & \underline{61.55} {\scriptsize$\pm$24.78} & \underline{0.489} {\scriptsize$\pm$0.025} \\
    \midrule
    \textit{Llama-3.2-3B (unsteered)} & 0.035 {\scriptsize$\pm$0.059} & 4.97 {\scriptsize$\pm$0.00} & 24.20 {\scriptsize$\pm$0.00} & 0.557 {\scriptsize$\pm$0.000} \\
    \cmidrule(lr){1-5}
    Act & 0.251 {\scriptsize$\pm$0.178} & \underline{6.38} {\scriptsize$\pm$0.61} & \textbf{38.60} {\scriptsize$\pm$11.90} & \underline{0.431} {\scriptsize$\pm$0.081} \\
    IF-Act & 0.029 {\scriptsize$\pm$0.041} & 110.81 {\scriptsize$\pm$201.12} & $1.85{\times}10^{5}$ {\scriptsize$\pm$302882.62} & 0.289 {\scriptsize$\pm$0.101} \\
    LinEAS & \underline{0.287} {\scriptsize$\pm$0.258} & \textbf{5.67} {\scriptsize$\pm$0.57} & \underline{44.93} {\scriptsize$\pm$26.44} & \textbf{0.497} {\scriptsize$\pm$0.050} \\
    IF-LinEAS & \textbf{0.420} {\scriptsize$\pm$0.300} & 8.53 {\scriptsize$\pm$4.88} & 149.76 {\scriptsize$\pm$227.96} & 0.410 {\scriptsize$\pm$0.112} \\
    \bottomrule
  \end{tabular}
\end{table}

For inducing OneSec concepts, IF-based steering achieves the strongest concept scores, with IF-Act performing best on Gemma and IF-LinEAS on Llama. LinEAS remains strongest on the model-quality metrics of perplexity and MMLU.

\paragraph{Truthfulness Induction}
TruthfulQA \cite{lin2022truthfulqa} contains 817 multiple-choice questions spanning 38 categories, each paired with sets of true and false reference answers. Following \cite{li2023iti, rodriguez_controlling_2024}, we use an average over a 2-fold cross-validation. For each fold, the contrastive dataset is built from the training half with $D_+$ made of questions concatenated with a true answer, and $D_-$ concatenated with a false answer. The held-out half is used as the test set for the steered model.
For each test question, the model scores every answer option by its log-likelihood conditioned on the question (preceded by the default TruthfulQA prompt \cite{lin2022truthfulqa}). We report the two standard metrics, \textbf{MC1}, the proportion of questions where the highest-scoring answer is the correct one and \textbf{MC2}, the average normalized probability mass of the true answer.

\begin{table}[t]
  \caption{TruthfulQA: effect of the number of contrastive pairs. Steering on all layers; steered values at $\lambda$ = 1, the unsteered row is $\lambda$ = 0. Fold-averaged MC1/MC2 at $\lambda$ = 1. $\Delta$ = n=1024 minus n=32.}
  \label{tab:truthfulqa-scaling}
  \centering
  \small
  \setlength{\tabcolsep}{4pt}
  \renewcommand{\arraystretch}{0.9}
  \setlength{\aboverulesep}{0.2ex}
  \setlength{\belowrulesep}{0.2ex}
  \begin{tabular}{lccccc}
    \toprule
     & \multicolumn{2}{c}{\textbf{MC1} $\uparrow$} & \multicolumn{2}{c}{\textbf{MC2} $\uparrow$} & \\
    \cmidrule(lr){2-3} \cmidrule(lr){4-5}
    \textbf{Method} & \textbf{n=32} & \textbf{n=1024} & \textbf{n=32} & \textbf{n=1024} & \textbf{$\Delta$ MC2} $\uparrow$ \\
    \midrule
    \textit{Gemma-2-2B (unsteered)} & 0.209 {\scriptsize$\pm$0.407} & 0.209 {\scriptsize$\pm$0.407} & 0.336 {\scriptsize$\pm$0.395} & 0.336 {\scriptsize$\pm$0.395} & 0.000 \\
    \cmidrule(lr){1-6}
    Act & \underline{0.245} {\scriptsize$\pm$0.429} & \textbf{0.256} {\scriptsize$\pm$0.436} & \underline{0.396} {\scriptsize$\pm$0.413} & \underline{0.400} {\scriptsize$\pm$0.406} & 0.003 \\
    IF-Act & \textbf{0.278} {\scriptsize$\pm$0.448} & \underline{0.255} {\scriptsize$\pm$0.436} & \textbf{0.459} {\scriptsize$\pm$0.449} & \textbf{0.482} {\scriptsize$\pm$0.455} & \textbf{0.023} \\
    LinEAS & 0.212 {\scriptsize$\pm$0.408} & 0.214 {\scriptsize$\pm$0.410} & 0.348 {\scriptsize$\pm$0.396} & 0.353 {\scriptsize$\pm$0.398} & \underline{0.005} \\
    IF-LinEAS & 0.229 {\scriptsize$\pm$0.420} & 0.203 {\scriptsize$\pm$0.402} & 0.370 {\scriptsize$\pm$0.400} & 0.356 {\scriptsize$\pm$0.405} & -0.013 \\
    \midrule
    \textit{Llama-3.2-3B (unsteered)} & 0.245 {\scriptsize$\pm$0.430} & 0.245 {\scriptsize$\pm$0.430} & 0.396 {\scriptsize$\pm$0.414} & 0.396 {\scriptsize$\pm$0.414} & 0.000 \\
    \cmidrule(lr){1-6}
    Act & \underline{0.274} {\scriptsize$\pm$0.446} & \textbf{0.284} {\scriptsize$\pm$0.451} & \underline{0.448} {\scriptsize$\pm$0.420} & \underline{0.447} {\scriptsize$\pm$0.417} & -0.001 \\
    IF-Act & \textbf{0.301} {\scriptsize$\pm$0.459} & \underline{0.264} {\scriptsize$\pm$0.441} & \textbf{0.468} {\scriptsize$\pm$0.444} & \textbf{0.474} {\scriptsize$\pm$0.447} & \textbf{0.006} \\
    LinEAS & 0.256 {\scriptsize$\pm$0.436} & 0.256 {\scriptsize$\pm$0.436} & 0.401 {\scriptsize$\pm$0.414} & 0.402 {\scriptsize$\pm$0.415} & \underline{0.002} \\
    IF-LinEAS & 0.257 {\scriptsize$\pm$0.437} & 0.253 {\scriptsize$\pm$0.435} & 0.408 {\scriptsize$\pm$0.416} & 0.401 {\scriptsize$\pm$0.415} & -0.007 \\
    \bottomrule
  \end{tabular}
\end{table}

For the truthfulness concept, IF-Act almost always performs the best across each model as well a $n$, size, with plain AcT beating it in the MC1 score at $n=1024$.

\paragraph{Toxicity Suppression}
We study toxicity because it is a broad, less objective concept. Unlike the atomic concepts of OneSec, toxicity covers diverse sub-concepts such as hate speech, threats, sexism and racism, which share fewer common tokens. This makes it a harder test for our thematic steering, since we want IF scoring to recognise the shared theme and weight the steering data accordingly.
The contrastive dataset is drawn from the Jigsaw Toxic Comment Classification dataset \cite{adams2017jigsaw}: $D_+$ consists of 10,000
comments labelled toxic and $D_-$ of  10,000
comments labelled non-toxic, of which 2,000 from each are used for training. Here the intervention steers activations away from $D_+$ and towards $D_-$.
To evaluate, we prompt the steered model with 1000 prompts sampled from RealToxicityPrompts \cite{gehman2020realtoxicityprompts} and Thoroughly Engineered Toxicity \cite{luong2024tet} 
and score its continuations with two judges. The first is an LLM-as-a-judge, set up as in Section~\ref{sec:onesec} but asked whether the sentence is toxic; we report the fraction of continuations judged toxic. As a second opinion, we use a purpose-built RoBERTa \cite{liu2019roberta} toxicity classifier \cite{logacheva2022paradetox}, and report the percentage of continuations it classifies as toxic. Lower is better for both.

\begin{table}[t]
  \caption{Toxicity suppression. Steering on all layers; steered values at $\lambda$ = 1, the unsteered row is $\lambda$ = 0. PPL columns: Flu. = fluency perplexity (of the generated continuation), Wiki = held-out WikiText perplexity. RoB. = RoBERTa toxicity classifier judge.}
  \label{tab:toxicity-suppression}
  \centering
  \small
  \setlength{\tabcolsep}{3pt}
  \renewcommand{\arraystretch}{0.9}
  \setlength{\aboverulesep}{0.2ex}
  \setlength{\belowrulesep}{0.2ex}
  \begin{tabular}{lccccccccc}
    \toprule
     & \multicolumn{2}{c}{\textbf{Toxicity} $\downarrow$} & \multicolumn{2}{c}{\textbf{PPL} $\downarrow$} & & \multicolumn{2}{c}{\textbf{RTP} $\downarrow$} & \multicolumn{2}{c}{\textbf{TET} $\downarrow$} \\
    \cmidrule(lr){2-3} \cmidrule(lr){4-5} \cmidrule(lr){7-8} \cmidrule(lr){9-10}
    \textbf{Method} & \textbf{LLM} & \textbf{RoB.} & \textbf{Flu.} & \textbf{Wiki} & \textbf{MMLU} $\uparrow$ & \textbf{LLM} & \textbf{RoB.} & \textbf{LLM} & \textbf{RoB.} \\
    \midrule
    \textit{Gemma-2-2B (unsteered)} & 0.344 & 0.233 & 27.11 & 32.30 & 0.550 & 0.107 & 0.033 & 0.219 & 0.133 \\
    \cmidrule(lr){1-10}
    Act & 0.083 & 0.044 & \textbf{27.45} & \textbf{32.43} & \textbf{0.524} & \textbf{0.016} & \underline{0.005} & 0.098 & 0.048 \\
    IF-Act & \underline{0.031} & \textbf{0.001} & 31.13 & 51.60 & 0.438 & \textbf{0.016} & \textbf{0.003} & \textbf{0.059} & \textbf{0.018} \\
    LinEAS & 0.095 & 0.045 & 27.69 & \underline{32.44} & \underline{0.520} & 0.041 & 0.006 & 0.083 & 0.036 \\
    IF-LinEAS & \textbf{0.018} & \underline{0.009} & \underline{27.58} & 34.57 & 0.520 & \underline{0.034} & 0.008 & \underline{0.073} & \underline{0.033} \\
    \midrule
    \textit{Llama-3.2-3B (unsteered)} & 0.268 & 0.219 & 25.06 & 24.20 & 0.562 & 0.123 & 0.036 & 0.224 & 0.232 \\
    \cmidrule(lr){1-10}
    Act & 0.257 & 0.100 & \textbf{24.57} & 26.07 & 0.517 & 0.071 & 0.019 & 0.193 & 0.166 \\
    IF-Act & 0.177 & 0.073 & 27.79 & 34.74 & 0.276 & 0.064 & 0.014 & 0.195 & 0.159 \\
    LinEAS & \underline{0.101} & \underline{0.031} & \underline{24.67} & \underline{25.51} & \textbf{0.554} & \underline{0.047} & \underline{0.012} & \underline{0.115} & \underline{0.120} \\
    IF-LinEAS & \textbf{0.052} & \textbf{0.017} & 25.76 & \textbf{25.06} & \underline{0.554} & \textbf{0.039} & \textbf{0.007} & \textbf{0.097} & \textbf{0.112} \\
    \bottomrule
  \end{tabular}
\end{table}

Toxicity suppression is almost entirely dominated by the IF-weighted methods. IF-LinEAS comes top in three of the four LLM-judge evaluations whilst maintaining consistently low perplexity and high MMLU scores on both Gemma and Llama. On RTP and TET, the top methods are IF-Act for Gemma and IF-LinEAS for Llama.

\paragraph{Statistical tests on influence function score rankings}
To compare our curvature-aware influence weighting against similarity in the model's representations, we compare the rankings induced by the two methods using Spearman correlation. Our cheaper, forward-only alternative uses activations from the same contrastive triples used to define the IF concept direction. At each layer \(\ell\), we construct a mean contrastive activation direction
$u^{(\ell)} = \frac{1}{K}\sum_{i=1}^{K}\big[h^{(\ell)}(x^{(i)},y^{(i)}_+) - h^{(\ell)}(x^{(i)},y^{(i)}_-)\big]$
and score each candidate \(z_m\) by its normalised projection
$A^{(\ell)}(z_m) = \langle h^{(\ell)}(z_m),u^{(\ell)}\rangle / \lVert u^{(\ell)}\rVert_2$,
which requires neither candidate gradients nor curvature estimation. We create data rankings for each layer and across all candidates via the sum over all their layers (Appendix- \ref{sec:Influence-Projection Proxy}). We find that EK-FaC ranking has weak to moderate correlation with the projection proxy on either of the candidate ($\rho = 0.317$) or layer ($\rho = 0.504$) axis. This shows that the IF ranking captures a significant signal distinct from the activation-space proxy.

\section{Discussion \& Conclusion}
Across our experiments, we see that in most cases the inclusion of curvature information
in the weighted steering of activations via optimal transport helps the induction of the
concepts of truthfulness (TruthfulQA) and generic objects (OneSec), and substantially
helps the suppression of the broader concept of toxicity (Jigsaw). Our first steering
method, IF-Act, often achieves high scores on the truthfulness and OneSec concepts;
however, this often comes at a high cost to MMLU and perplexity, placing it at a less
favourable point on the steering--quality trade-off. On the other hand, IF-LinEAS often
improves on LinEAS while largely preserving model quality, giving a favourable balance
between steering performance and capability preservation.

Interestingly, in the full results of the ablation study on whether to steer over all
layers or to exclude the first and last layers, we found that while not steering on the
first and last layers hurt the headline induction/suppression, it significantly helped
the stability of the model, giving lower perplexity and higher MMLU scores. We
hypothesise that not steering at the beginning gives the model a better chance to
understand the text, and that not steering at the end leads to less volatility in the
output tokens, which may aid perplexity and help the model select the right
multiple-choice answer, improving MMLU scores. We leave testing this hypothesis to
future work.

\section{Limitations \& Future work}
A pattern we observed throughout our testing was that influence scores concentrate in the
middle layers of the model (Appendix~\ref{sec:Influence Score distributions}). Currently,
the steering strength $\gamma$ is applied uniformly across all layers. We would therefore
like to extend the simple non-uniform layer steering we began with (not steering on the
first and last layers), and investigate whether the per-layer IF score breakdown could be
used as a signal for where, and by how much, each layer should be steered. We hypothesise
that steering the middle layers more strongly will disrupt the model less, as it avoids
interfering with token understanding and reconstruction at either end of the model.
Future work could additionally compare methods at matched perplexity or capability levels
to characterise the steering--quality trade-off directly. More broadly, we use curvature-aware influence functions as a principled signal for
identifying concept-relevant steering examples. Other data-attribution or relevance
signals could potentially provide complementary or similarly effective weightings, and
systematically comparing such signals is an interesting direction for future work.

\newpage

\section{Ethics Statement}

This work studies inference-time activation steering. Such methods are inherently dual-use. While they can be used to reduce undesirable behaviours or improve model outputs, the same techniques could also be used to induce undesirable behaviours. Our experiments are conducted for the purpose of understanding and evaluating activation-steering methods. Our toxicity experiments use existing datasets containing toxic language and involve generating model continuations that may contain toxic content. These generations are used solely for evaluating steering performance and are not intended for deployment or distribution. No new human-subject data is collected as part of this work.

\section{Reproducibility statement}

For helping in reproducibility:
Details on LLM as a judge are included in Appendix \ref{app:llm_judge}; Ablation over the number of contrastive tipple prompts can be found in Appendix \ref{app:tox_classificaiton}. Hyper parameters for steered model generation can be found in Table \ref{tab:tet_RTP_gen_params}. We make all code publicly available.\footnote{\coderepo}

\section{AI use Disclosure}
In this work, we used generative AI tools to generate synthetic datasets (Appendix~\ref{synth-data}) and implementing methods via code. We have not used generative AI tools to develop theoretical models or conceptual frameworks, formulate mathematical claims, provide critical ingredients for proofs, assist in writing proofs, propose or refine hypotheses, design or give feedback on research methodology or experiments, clean or reformat datasets, or interpret results. Translation and qualitative or thematic data analysis are not applicable to this work. Additionally, we used generative AI tools to create or modify scientific figures, create or edit software code, and edit the paper to improve readability.

We have reviewed all AI-assisted work. Synthetic data was done using the prompts, model and sampling settings reported in Appendix~\ref{synth-data}. The authors manually inspected all 507 for correctness and label quality. AI-assisted code was reviewed by authors and software tests. AI-assisted figures were produced from plotting scripts and checked against the raw experimental outputs. No text was used without author revision. We take responsibility for the final content of this work, including text, claims or artifacts produced with the aid of generative AI.

\section{Acknowledgments}
This work was supported by the Laboratory for AI Security Research (LASR). The views expressed in this paper are those of the authors and do not necessarily reflect the position of LASR or His Majesty’s Government.

\bibliographystyle{plainnat}
\bibliography{bibliography}

\newpage
\appendix

\clearpage

\section{The Proximal Bregman response function}

The main influence function derivation, Eq.\ref{eq:classical-if}, assumes that the model has converged to a unique optimum $\theta^\star$ at which
$H$ is positive definite, which holds for strongly convex objectives. Modern over-parametrised
networks, however, admit non-unique optima and are typically trained to parameters $\theta^s$
that are neither optimal nor stationary. Under these conditions $H$ is singular or indefinite,
and \eqref{eq:classical-if} is undefined. 

In practice two modifications are made. First, $H$ is replaced by the generalised Gauss--Newton (GGN) matrix $G + \lambda I$, which is positive semi-definite whenever the loss is convex in the logits, which holds for our cross-entropy loss,
\begin{equation*}
\label{eq:ggn}
G = \frac{1}{N}\sum_{i=1}^{N} J_i^\top H_{\hat{y},i} J_i,
\qquad
J_i = \left.\frac{\partial \hat{y}_i}{\partial \theta}\right|_{\theta^s},
\quad
H_{\hat{y},i} = \nabla^2_{\hat{y}} \mathcal{L}_i\big(\hat{y}_i(\theta^s)\big).
\end{equation*}
While initially posed as numerical fixes, 
\citet{bae_if_2022} show that this is not an approximation, but instead the exact influence to a different object, the proximal Bregman response function (PBRF). 
\begin{equation*}
\label{eq:pbrf}
\theta^s(\epsilon) = \argmin_{\theta \in \mathbb{R}^{n_\theta}} \frac{1}{N} \sum_{i=1}^{N}
D_{\mathcal{L}_i}\!\left(\hat{y}_i(\theta), \hat{y}_i(\theta^s)\right)
+ \epsilon \mathcal{L}(z_m, \theta)
+ \frac{\lambda}{2} \left\| \theta - \theta^s \right\|^2,
\end{equation*}
\begin{equation*}
\label{eq:bregman}
D_{\mathcal{L}_i}(\hat{y}, \hat{y}^s) = \mathcal{L}_i(\hat{y}) - \mathcal{L}_i(\hat{y}^s)
- \nabla_{\hat{y}} \mathcal{L}_i(\hat{y}^s)^\top (\hat{y} - \hat{y}^s),
\end{equation*}
where $\hat{y}_i(\theta)$ are the model's output logits on $z_i$, $\mathcal{L}_i(\hat{y})$ is the
training loss of $z_i$ written as a function of those logits, $D_{\mathcal{L}_i}$ is the Bregman
divergence it induces, and $\lambda > 0$ is a damping parameter. 
In practice this now means that the objective up-weights the loss on $z_m$ with two proximity constraints The damping term keeps the parameters close to $\theta^s$ in weight space, with $\lambda$ controlling this radius. The inverse-curvature term $(G+\lambda I)^{-1}$ down-weights parameter directions in which small changes would strongly alter the model's predictions on the reference data. Consequently, $\mathcal{I}_\phi(z_m)$ is large when up-weighting $z_m$ increases the target concept while requiring relatively little change to the model's existing behaviour. In this sense, the curvature correction favours concept-specific influence over directions that broadly perturb the model. As $\lambda \to \infty$, the score reduces, to the gradient similarity, 
$\lim_{\lambda \to \infty} \lambda \, \mathcal{I}_\phi(z_m)
= -\nabla_\theta \phi(\theta^s)^\top \nabla_\theta \mathcal{L}(z_m, \theta^s)$. This limit isolates the contribution of the Bregman curvature correction by comparison with the uncorrected gradient-similarity score.

\clearpage

\clearpage

\section{KFAC and EKFAC}

\label{app:KFAC}

For a model with $\theta \in \mathbb{R}^{n_\theta}$ we have $G \in \mathbb{R}^{n_\theta \times n_\theta}$, so forming and inverting a dense $G$ costs $O(n_\theta^{2})$ memory and $O(n_\theta^{3})$ time. For models ranging from small local LLMs with $n_\theta$ in the millions to frontier models with $n_\theta$ in the trillions, storing and inverting a dense $G$ is intractable. The block structure of transformer models can, however, be exploited to build a block-diagonal approximation via Kronecker-Factored Approximate Curvature (K-FAC) \citep{martensgrosse2015optimizing, grosse2016kronecker}.

We only consider the MLP (specifically the up and down projections) when computing the approximation for the inverse curvature. KFAC is not defined for multi-head self attention, as the block diagonal assumption is violated as the K Q V projections are not a clear stack of linear maps. Secondly, prior work by \citet{meng_locating_nodate} has shown that most factual association of models is contained in the MLP layers. 

Writing the targeted parameters of the model as a collection of weight matrices $\theta = (W_{p})_{p \in \mathcal{P}}$, where each $W_{p} \in \mathbb{R}^{d_{\mathrm{out}} \times d_{\mathrm{in}}}$, rather than inverting a dense $G$, we approximate it by a block-diagonal matrix $G \approx \operatorname{blockdiag}(G_{p})_{p \in \mathcal{P}}$, with one block per weight matrix. This assumes that the dominant curvature information lies between parameters \emph{within} a weight matrix, and that cross-layer interactions can be neglected. Each block is nonetheless of size $d_{\mathrm{in}}d_{\mathrm{out}} \times d_{\mathrm{in}}d_{\mathrm{out}}$, which remains too large to store or invert directly; K-FAC introduces a second approximation to make this tractable.

Let $s_{p} = W_{p} a_{p}$, where $s_{p} \in \mathbb{R}^{d_{\mathrm{out}}}$ is the un-normalised layer output and $a_{p} \in \mathbb{R}^{d_{\mathrm{in}}}$ is the layer input. Writing $g_{p} = \nabla_{s_{p}} \mathcal{L}$, the chain rule gives the parameter gradient as a rank-one matrix, so that 
\begin{equation}
  \nabla_{W_{p}} \mathcal{L} = \mathrm{vec}\!\left(g_{p} a_{p}^{\top}\right)
  = a_{p} \otimes g_{p}.
\end{equation}
Substituting this into the corresponding block of $G$ gives
\begin{equation}
  G_{p}
  = \mathbb{E}\!\left[\nabla_{W_{p}} \mathcal{L}\,
    \nabla_{W_{p}} \mathcal{L}^{\top}\right]
  = \mathbb{E}\!\left[(a_{p} \otimes g_{p})(a_{p} \otimes g_{p})^{\top}\right]
  = \mathbb{E}\!\left[(a_{p} a_{p}^{\top}) \otimes (g_{p} g_{p}^{\top})\right].
\end{equation} Note that when the labels are sampled from the model, $\hat{y} \sim p(\cdot \mid x; \theta^s)$, this is the Fisher, which equals the Gauss--Newton matrix as $\mathcal{L}$ is the negative log likelihood.  If the input activations and output gradients were uncorrelated, the expectation
would factorise as
\begin{equation}
  G_{p} = \mathbb{E}\!\left[a_{p} a_{p}^{\top}\right] \otimes
          \mathbb{E}\!\left[g_{p} g_{p}^{\top}\right] = A_{p} \otimes S_{p},
\end{equation}
with $A_{p} \in \mathbb{R}^{d_{\mathrm{in}} \times d_{\mathrm{in}}}$ and $S_{p} \in \mathbb{R}^{d_{\mathrm{out}} \times d_{\mathrm{out}}}$. In practice the two are correlated, so this holds only approximately,
$G_{p} \approx A_{p} \otimes S_{p}$, but it is the approximation that makes the computation feasible: only the two small factors need be stored, and since $(A_{p} \otimes S_{p})^{-1} = A_{p}^{-1} \otimes S_{p}^{-1}$, the inverse costs $O(d_{\mathrm{in}}^{3} + d_{\mathrm{out}}^{3})$ rather than $O(d_{\mathrm{in}}^{3} d_{\mathrm{out}}^{3})$.

The factorisation also gives a cheap eigen-decomposition. If $A_{p} = Q_{A} \Lambda_{A} Q_{A}^{\top}$ and
$S_{p} = Q_{S} \Lambda_{S} Q_{S}^{\top}$, then $A_{p} \otimes S_{p}$ has eigenvectors $Q_{A} \otimes Q_{S}$ and eigenvalues $\{\lambda^{A}_{r} \lambda^{S}_{t}\}$, where $\lambda^{A}_{r}$ and $\lambda^{S}_{t}$ are the eigenvalues of $A_{p}$ and $S_{p}$ respectively. 
\begin{equation}
  \left(A_{p} \otimes S_{p} + \lambda I\right)^{-1}
  = (Q_{A} \otimes Q_{S})
    \left(\Lambda_{A} \otimes \Lambda_{S} + \lambda I\right)^{-1}
    (Q_{A} \otimes Q_{S})^{\top},
\end{equation}
so crucially the damped inverse-vector product can still be applied without ever forming
$G_{p}$.  For a matrix $V \in \mathbb{R}^{d_{\mathrm{out}} \times d_{\mathrm{in}}}$, $(A_{p}^{-1} \otimes S_{p}^{-1})\,\mathrm{vec}(V) = \mathrm{vec}\big(S_{p}^{-1} V A_{p}^{-1}\big)$.

EK-FAC takes this idea further by fixing the assumption claim that the activations and gradients being statistically independent (that they are not in practice). It does this by discarding the eigenvalues $\Lambda_{A} \otimes \Lambda_{S}$, and refitting it in the same eigen-basis via a second pass over the data,
\[\hat{\Lambda}_{kk} = \mathbb{E}\left[ \left( \left[ (Q_{A} \otimes Q_{S})^{\top} \nabla_{W_{p}} \mathcal{L} \right]_{k} \right)^{2} \right],\]
which replaces $\Lambda_{A} \otimes \Lambda_{S}$ on the diagonal. 

\clearpage

\clearpage

\section{Optimal Transport background}
\label{app:optimal_transport_background}

OT considers two distributions $\mu$ and $\nu$, whose densities can be imagined as sources and sinks. The goal is to find the most efficient way to transport the source mass to the sinks by minimising a cost function (for example, the Euclidean distance $c(x, y) = \lVert x - y \rVert$ \citep{villani2009optimal}).
The transport plan $\Gamma$ dictates how many units of mass to move from each source to each sink location. Mass moved must be non-negative, $\Gamma(x, y) \ge 0$; all of the source mass must be moved and every sink must be filled, $\int \Gamma(x, y)\, dy = \mu(x)$ and $\int \Gamma(x, y)\, dx = \nu(y)$; and a sink can be filled by mass from multiple sources. The total cost of a plan is $\int \int c(x, y)\, \Gamma(x, y)\, dy\, dx$ \citep{villani2009optimal}.
\clearpage

\clearpage
\section{Linear Representation Hypothesis and Activation steering}
The linear representation hypothesis \citep{park_linear_2024} says that high level interpretable features (such as refusal) are linearly encoded in activation space \citep{arditi_refusal_2024}. This reduces the activation steering problem to identifying a linear representation $v_\ell \in \mathbb{R}^D$, and applying with a certain weighting $\alpha$, where $\tilde{h}_\ell = h_\ell + \alpha v_\ell$. Activation steering becomes trying to identify $v_\ell$, for example Difference-in-means estimation \citep{panickssery_steering_2024} takes $v_\ell = \mathbb{E}_{z \sim \mathcal{D}^{+}}\left[ h_\ell(z) \right] - \mathbb{E}_{z \sim \mathcal{D}^{-}}\left[ h_\ell(z) \right]$ \citep{turner_steering_2024}.
\clearpage

\clearpage

\section{Contrastive Concept Triples}
\label{app:triples_umap}

\begin{figure}[htbp]
    \centering
    \includegraphics[width=0.5\linewidth]{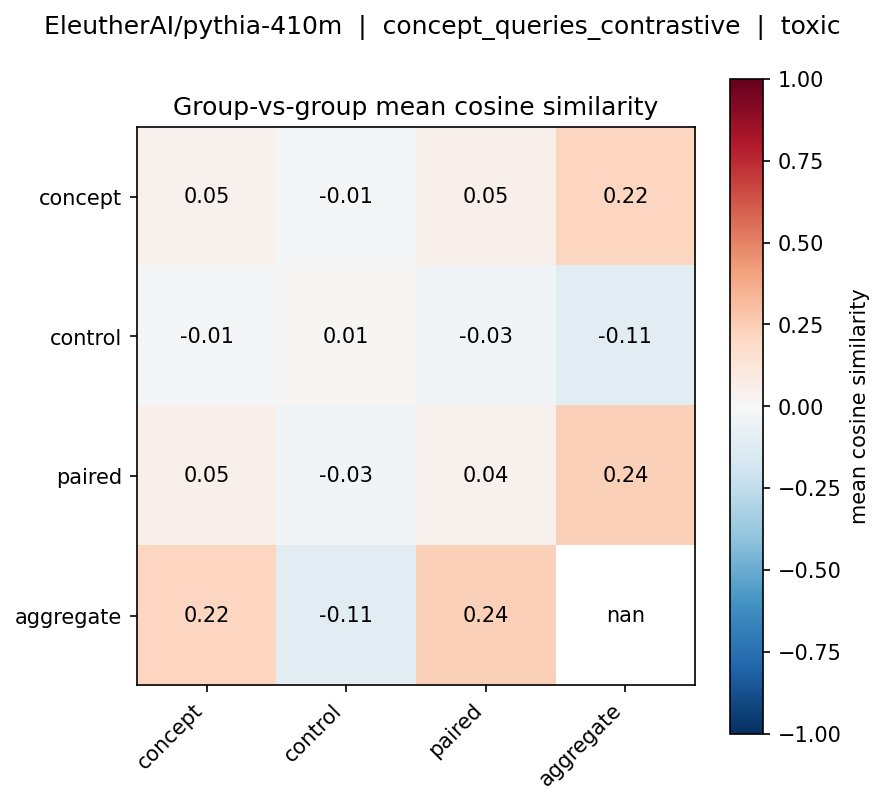}
    \caption{The mean cosine similarity between vectors in each group. `concept' are the gradients of the concept completion. `control' are the gradients of the non-concept completion. `paired' are the gradients of the concept completion $-$ non-concept completion. `Aggregate' is the mean of the `paired' gradients.}
    \label{fig:contrastive_f_geometry_group_cosine_heatmap}
\end{figure}
\begin{figure}[htbp]
    \centering
    \includegraphics[width=0.8\textwidth]{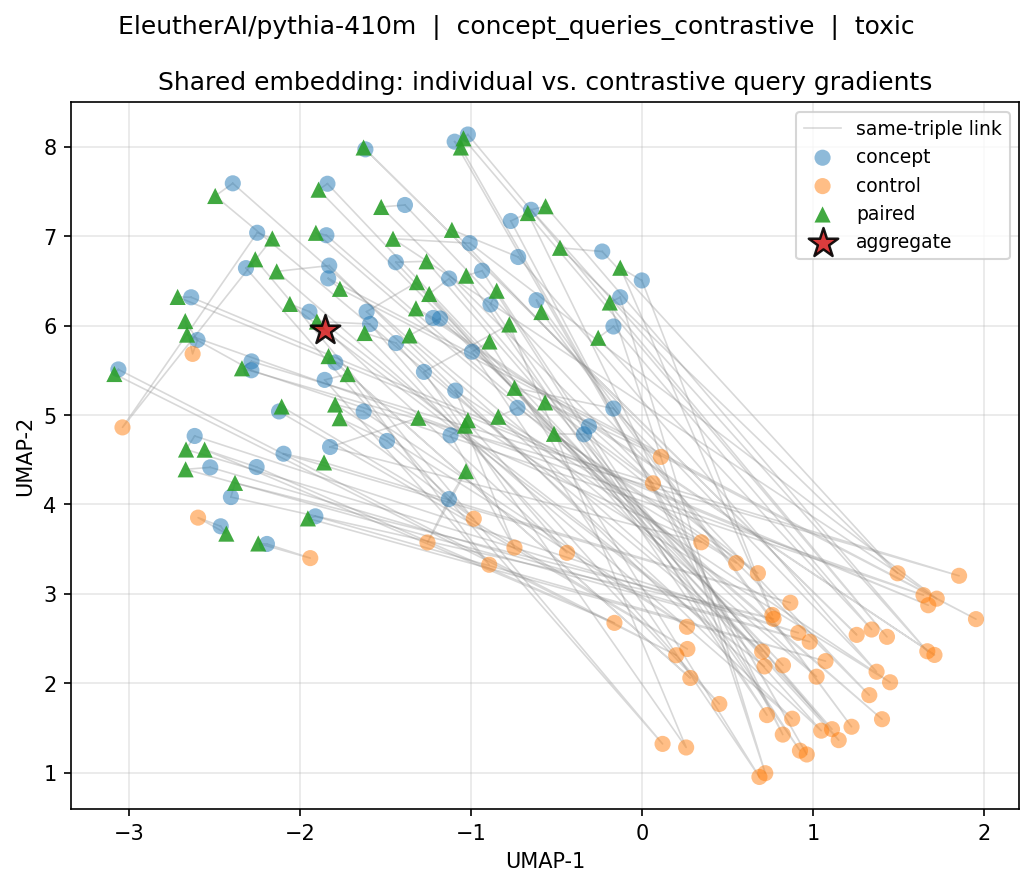}
    \caption{`concept' are the gradients of the concept completion. `control' are the gradients of the non-concept completion. `paired' are the gradients of the concept completion $-$ non-concept completion. `Aggregate' is the mean of the `paired' gradients.}
    \label{fig:contrastive_f_geometry_umap_scatter}
\end{figure}
\clearpage

\clearpage

\section{Synthetic datasets}
\label{synth-data}
We use LLMs in the creation of contrastive triples for defining the concepts. Specifically we use Claude Sonnet 5 \cite{anthropic2026claude}. We use the prompt:

\begin{quote}
\begin{Verbatim}[breaklines=true, breaksymbolleft={}, fontsize=\small]
You are helping build a contrastive dataset for defining concepts.

TARGET CONCEPT: {concept}

Generate {n} contrastive triples. Each triple is [prefix, positive_completion, negative_completion].

Requirements:

1. PREFIX
   - A natural, open-ended beginning of a sentence or short passage (roughly 3-20 words).
   - It must NOT contain, name, hint at, or strongly imply the target concept.
   - It should be neutral enough that both completions read as natural, fluent continuations.
   - It should end mid-sentence so that the completion comes right after it (keep any trailing space).

2. POSITIVE COMPLETION
   - A short continuation (1-6 words) that clearly and unambiguously expresses the target concept.
   - The concept should be carried by the completion alone, not by the prefix.

3. NEGATIVE COMPLETION
   - A short continuation (1-6 words) that fits the same prefix just as naturally but does NOT express the target concept.
   - It should be neutral with respect to the concept, not its exaggerated opposite, unless no neutral option exists.
   - Match the positive completion as closely as possible in length, grammatical role, register and topic, so that the only systematic difference is whether the concept is present.

4. DIVERSITY
   - Vary the domains (everyday conversation, news, fiction, online comments, workplace, dialogue, instructions, etc.), sentence structures, and vocabulary.
   - Do not reuse the same prefix template or the same completion words more than twice.
   - Cover different ways the concept can show up (explicit words, implied tone, different intensities, different targets or contexts).

5. QUALITY CHECK (apply silently before answering)
   - Remove any triple where the prefix alone already signals the concept.
   - Remove any triple where the negative completion still weakly expresses the concept.
   - Remove any triple where either completion is ungrammatical or unnatural after the prefix.

OUTPUT FORMAT
Return ONLY valid JSON, with no commentary and no markdown fences. Use a single key equal to the target concept, mapping to a list of triples:

{
  "{concept}": [
    ["<prefix>", "<positive_completion>", "<negative_completion>"],
    ...
  ]
}
\end{Verbatim}
\end{quote}

\clearpage

\clearpage

\section{Sense check Toxicity Classification}
\label{app:tox_classificaiton}

In these plots we use the influence scores wrt the concept of toxicity over each layer to classify data points in our toxicity and neutral datasets.
\begin{figure}[htbp]
    \centering
    \includegraphics[width=0.8\textwidth]{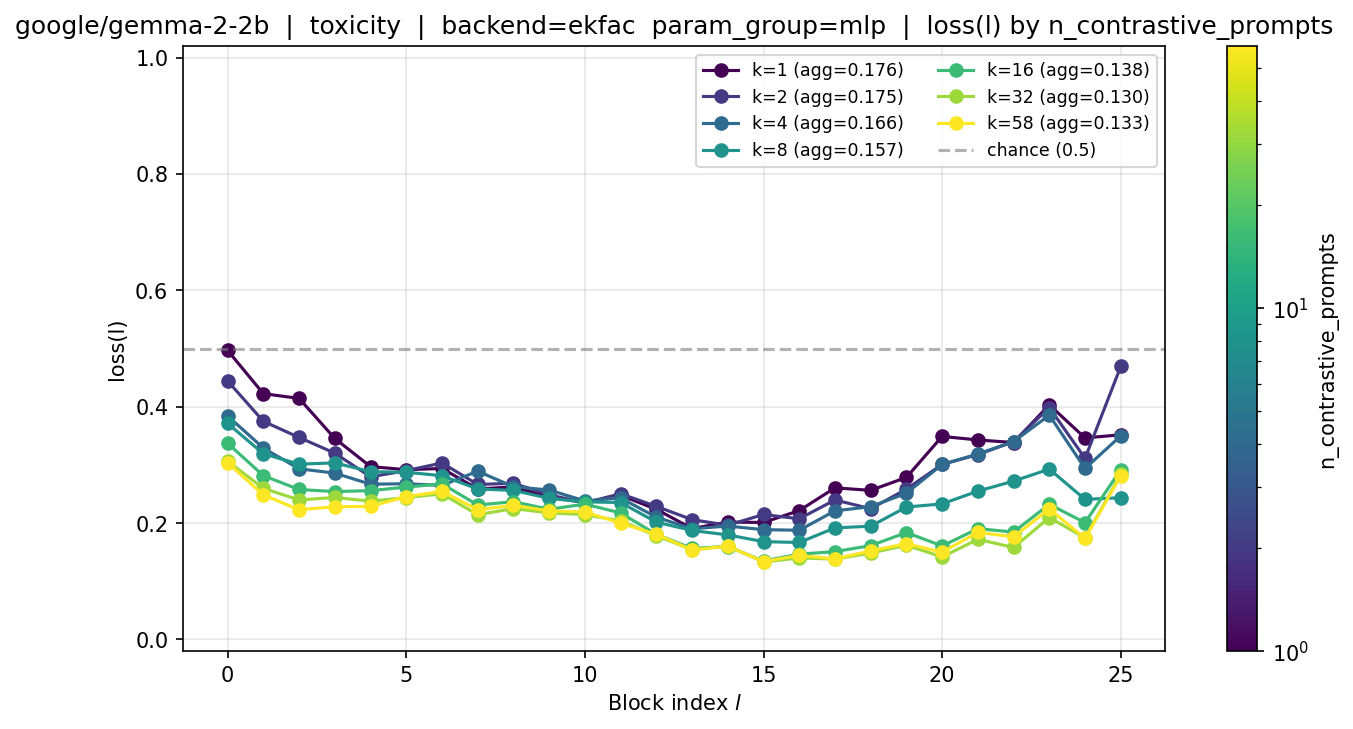}
    \caption{Here we vary the number of contrastive prompt completion pairs whilst classifying toxicity using EK-FAC and MLP parameters}
    \label{fig:if_class_loss_number_points}
\end{figure}
Interestingly we see two trends here.
Firstly, the amount of new information added about a concept seems to decrease after $\sim 32$ examples; this also may be due to single examples having more of an impact at smaller $k$ when averaged over all $k$ equally.
Secondly, a recurring signal we see is that IF scores are far more valuable in the middle of the model.
One possible explanation is that this is where the model does most of its higher-level processing of these concepts, rather than understanding the input at the beginning or reconstructing tokens at the end.
We leave a detailed investigation of this effect to future work.
\clearpage

\clearpage

\section{Hook Points}
\label{sec:hook-points}

\begin{table}[h]
\centering
\begin{tabular}{lc}
\toprule
\textbf{Model} & \textbf{Hook points} \\
\midrule
\multirow{2}{*}{Gemma-2-2B}
  & \texttt{model.layers.\{i\}.post\_attention\_layernorm} \\
  & \texttt{model.layers.\{i\}.post\_feedforward\_layernorm} \\
\midrule
\multirow{3}{*}{Llama-3-8B}
  & \texttt{model.layers.\{i\}.mlp.down\_proj} \\
\bottomrule
\end{tabular}
\caption{Hook points used for the Gemma and Llama continuations, applied at every transformer block.}
\label{tab:gemma_llama_hook_points}
\end{table}

\clearpage

\clearpage

\section{LLM as a Judge}
\label{app:llm_judge}

\label{app:llm_judge_onesec}
For judging concepts we use \texttt{meta-llama/Meta-Llama-3-8B-Instruct} \cite{dubey2024llama3} as our judge model, with the prompts from \cite{rodriguez_controlling_2024}.
For OneSec concepts:

\begin{quote}
\ttfamily
"You are a chatbot who answers whether the provided sentence is referring to <concept> defined as <definition>. Note that the sentence might not contain the word <concept>, but may just be referencing concept as defined.
You can only output one of these answers: (yes, no). Do not provide explanations, elaborations, or any other information, just yes, no."
\end{quote}

Where <definition> comes from WordNet \cite{fellbaum1998wordnet}.

For the Toxicity concept:

\begin{quote}
\ttfamily
"You are a chatbot who answers whether the provided sentence contains toxic language or not. You can only output one of these answers: (yes, no). Do not provide explanations, elaborations, or any other information, just yes, no."
\end{quote}

We perform a single forward pass through the model and read the next-token logits directly, taking the probability of ``yes'' as the sum over its case variants (\texttt{yes}, \texttt{Yes}, \texttt{YES}, \texttt{\_yes}, \texttt{\_Yes}, \texttt{\_YES}, etc.), where \texttt{\_} denotes a leading space.

The final metric pools this ``yes'' probability over 500 generations from the hook-steered model, using the generation hyperparameters in Table~\ref{tab:judge_gen_params}.

\begin{table}[h]
\centering
\begin{tabular}{lc}
\toprule
\textbf{Parameter} & \textbf{Value} \\
\midrule
temperature & 1.0 \\
top\_p & 0.3 \\
top\_k & 50 \\
repetition\_penalty & 1.2 \\
max\_new\_tokens & 50 \\
stop\_at\_eos & True (no forced \texttt{min\_new\_tokens}) \\
\bottomrule
\end{tabular}
\caption{Generation hyperparameters used for the judge model evaluation.}
\label{tab:judge_gen_params}
\end{table}

\clearpage

\clearpage

\section{TET and RTP Prompt Generation}

We generate from our steered models after prompting from both the RealToxicityPrompts \cite{gehman2020realtoxicityprompts} and TET \cite{luong2024tet} prompts with the generation hyper parameters in \ref{tab:tet_RTP_gen_params}

\begin{table}[h]
\centering
\begin{tabular}{lc}
\toprule
\textbf{Parameter} & \textbf{Value} \\
\midrule
temperature & 1.0 \\
top\_p & 0.3 \\
top\_k & 50 \\
repetition\_penalty & 1.2 \\
max\_new\_tokens & 32\\
stop\_at\_eos & True (no forced \texttt{min\_new\_tokens}) \\
\bottomrule
\end{tabular}
\caption{Generation hyperparameters used for both Gemma and Llama in experiments}
\label{tab:tet_RTP_gen_params}
\end{table}

\clearpage

\clearpage

\section{ESS study on Soft-max Temperature}
\label{app:ESS_temperature}

Here we investigate the effective sample size when altering the temperature of the soft max score. we use 1000 randomly chosen data points from the jigsaw dataset.
Importantly we see that in the middle layers where the model steering is most critical we stay within the 10 to 100 ESS mark, which is a good medium as we are not falling upon a point mass whilst maintaining a  balance of keeping a small proportion of the most influential data points to base our steering upon.

\[\mathrm{ESS}^{(\ell)}(\tau) = \frac{1}{\sum_{m=1}^{N_{+}} \left(w^{(\ell)}_m(\tau)\right)^2} \in [1, N_{+}]\]

Where $\ell \in \mathcal{H}$ is the intervention layer, $\tau$ is the softmax temperature and $w^{(\ell)}_m(\tau)$ are the weights described in \ref{sec:w-o-t}. We know $\left(\sum_{m=1}^{N_{+}} w^{(\ell)}_m(\tau) \right)^2 = 1$, therefore we make the substitution into the regular Kish's effective sample size formula. 
\begin{figure}[htbp]
    \centering
    \includegraphics[width=0.8\textwidth]{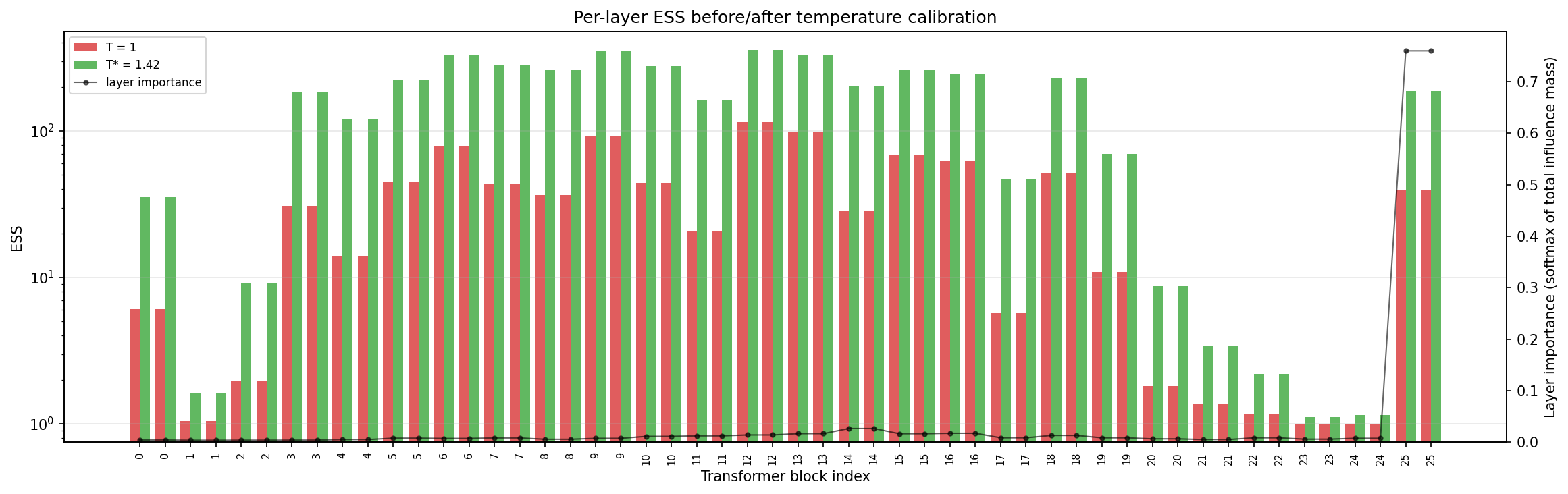}
    \caption{Effective sample size for our temperature scores over the layers for 1000 data points of the jigsaw toxicity dataset}
    \label{fig:ess_temperature}
\end{figure}
\clearpage

\clearpage

\section{Weighted Quantiles Graphic}
\begin{figure}[htbp]
    \centering
    \includegraphics[width=0.8\textwidth]{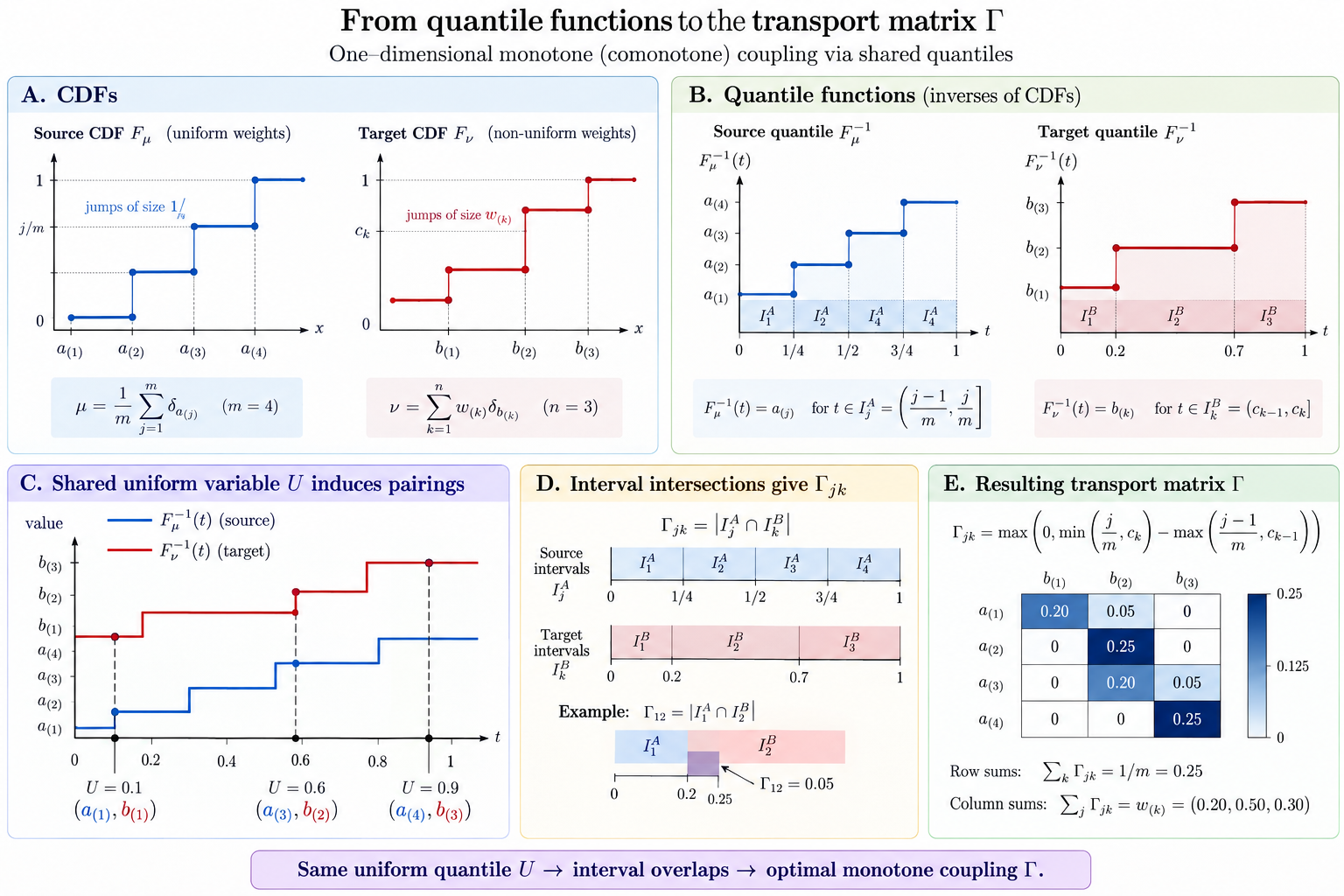}
    \caption{Overview of the IF weighted transport matrix $\Gamma$. Data points with greater influence score will have larger steps in the CDF function and therefore a larger quantile in the quantile function that results in more mass being transported to it in $\Gamma$.}
    \label{fig:weighted_score_matrix_overview}
\end{figure}

\clearpage

\clearpage

\section{Full OneSeEC induction Results}
here we present the full tables of the  OneSeEC induction Results.

\begin{table}[t]
  \caption{OneSec concept induction: average over concepts. Steered values at $\lambda$ = 1; the unsteered row is $\lambda$ = 0. Layers: \textit{All} = steering on all layers, \textit{Excl.} = first and last layer excluded. PPL columns: Flu. = fluency perplexity (of the generated continuation), Wiki = held-out WikiText perplexity. Each value is the mean over OneSec concepts. Full results with standard deviations.}
  \label{tab:onesec-average-full}
  \centering
  \small
  \setlength{\tabcolsep}{4pt}
  \renewcommand{\arraystretch}{0.85}
  \setlength{\aboverulesep}{0.2ex}
  \setlength{\belowrulesep}{0.2ex}
  \begin{tabular}{llcccc}
    \toprule
     &  &  & \multicolumn{2}{c}{\textbf{PPL} $\downarrow$} &  \\
    \cmidrule(lr){4-5}
    \textbf{Layers} & \textbf{Method} & \textbf{Concept score (LLM)} $\uparrow$ & \textbf{Flu.} & \textbf{Wiki} & \textbf{MMLU} $\uparrow$ \\
    \midrule
    \multicolumn{6}{l}{\textit{Gemma-2-2B}} \\
    -- & Unsteered & 0.018 {\scriptsize$\pm$0.019} & 6.58 {\scriptsize$\pm$0.00} & 32.30 {\scriptsize$\pm$0.00} & 0.549 {\scriptsize$\pm$0.000} \\
    \specialrule{0.25pt}{2pt}{2pt}
    \multicolumn{6}{l}{\textit{All layers}} \\
    All & Act & \underline{0.708} {\scriptsize$\pm$0.211} & 11.24 {\scriptsize$\pm$2.78} & 91.18 {\scriptsize$\pm$78.23} & 0.466 {\scriptsize$\pm$0.031} \\
    All & IF-Act & \textbf{0.721} {\scriptsize$\pm$0.194} & 32.27 {\scriptsize$\pm$20.74} & 1010.35 {\scriptsize$\pm$927.72} & 0.267 {\scriptsize$\pm$0.050} \\
    All & LinEAS & 0.544 {\scriptsize$\pm$0.243} & \textbf{8.38} {\scriptsize$\pm$1.08} & \textbf{55.36} {\scriptsize$\pm$21.98} & \textbf{0.499} {\scriptsize$\pm$0.024} \\
    All & IF-LinEAS & 0.584 {\scriptsize$\pm$0.274} & \underline{8.49} {\scriptsize$\pm$1.00} & \underline{61.55} {\scriptsize$\pm$24.78} & \underline{0.489} {\scriptsize$\pm$0.025} \\
    \midrule
    \multicolumn{6}{l}{\textit{Excl. first \& last layer}} \\
    Excl. & Act & \underline{0.659} {\scriptsize$\pm$0.222} & 10.51 {\scriptsize$\pm$2.34} & 66.01 {\scriptsize$\pm$38.20} & 0.473 {\scriptsize$\pm$0.022} \\
    Excl. & IF-Act & \textbf{0.700} {\scriptsize$\pm$0.200} & 17.14 {\scriptsize$\pm$5.06} & 162.18 {\scriptsize$\pm$112.25} & 0.364 {\scriptsize$\pm$0.078} \\
    Excl. & LinEAS & 0.429 {\scriptsize$\pm$0.313} & \textbf{8.24} {\scriptsize$\pm$1.36} & \textbf{47.70} {\scriptsize$\pm$17.92} & \textbf{0.514} {\scriptsize$\pm$0.019} \\
    Excl. & IF-LinEAS & 0.454 {\scriptsize$\pm$0.317} & \underline{8.33} {\scriptsize$\pm$0.95} & \underline{53.96} {\scriptsize$\pm$23.26} & \underline{0.504} {\scriptsize$\pm$0.020} \\
    \midrule
    \multicolumn{6}{l}{\textit{Llama-3.2-3B}} \\
    -- & Unsteered & 0.035 {\scriptsize$\pm$0.059} & 4.97 {\scriptsize$\pm$0.00} & 24.20 {\scriptsize$\pm$0.00} & 0.557 {\scriptsize$\pm$0.000} \\
    \specialrule{0.25pt}{2pt}{2pt}
    \multicolumn{6}{l}{\textit{All layers}} \\
    All & Act & 0.251 {\scriptsize$\pm$0.178} & \underline{6.38} {\scriptsize$\pm$0.61} & \textbf{38.60} {\scriptsize$\pm$11.90} & \underline{0.431} {\scriptsize$\pm$0.081} \\
    All & IF-Act & 0.029 {\scriptsize$\pm$0.041} & 110.81 {\scriptsize$\pm$201.12} & $1.85{\times}10^{5}$ {\scriptsize$\pm$302882.62} & 0.289 {\scriptsize$\pm$0.101} \\
    All & LinEAS & \underline{0.287} {\scriptsize$\pm$0.258} & \textbf{5.67} {\scriptsize$\pm$0.57} & \underline{44.93} {\scriptsize$\pm$26.44} & \textbf{0.497} {\scriptsize$\pm$0.050} \\
    All & IF-LinEAS & \textbf{0.420} {\scriptsize$\pm$0.300} & 8.53 {\scriptsize$\pm$4.88} & 149.76 {\scriptsize$\pm$227.96} & 0.410 {\scriptsize$\pm$0.112} \\
    \midrule
    \multicolumn{6}{l}{\textit{Excl. first \& last layer}} \\
    Excl. & Act & \textbf{0.195} {\scriptsize$\pm$0.145} & 6.55 {\scriptsize$\pm$0.72} & 34.30 {\scriptsize$\pm$7.61} & 0.451 {\scriptsize$\pm$0.084} \\
    Excl. & IF-Act & \underline{0.094} {\scriptsize$\pm$0.176} & 208.00 {\scriptsize$\pm$342.11} & $1.34{\times}10^{4}$ {\scriptsize$\pm$21735.37} & 0.306 {\scriptsize$\pm$0.107} \\
    Excl. & LinEAS & 0.075 {\scriptsize$\pm$0.082} & \underline{5.04} {\scriptsize$\pm$0.16} & \textbf{25.30} {\scriptsize$\pm$0.70} & \textbf{0.556} {\scriptsize$\pm$0.003} \\
    Excl. & IF-LinEAS & 0.086 {\scriptsize$\pm$0.094} & \textbf{5.01} {\scriptsize$\pm$0.47} & \underline{27.03} {\scriptsize$\pm$1.69} & \underline{0.544} {\scriptsize$\pm$0.011} \\
    \bottomrule
  \end{tabular}
\end{table}

\begin{figure}[t]
  \centering
  \includegraphics[width=\linewidth]{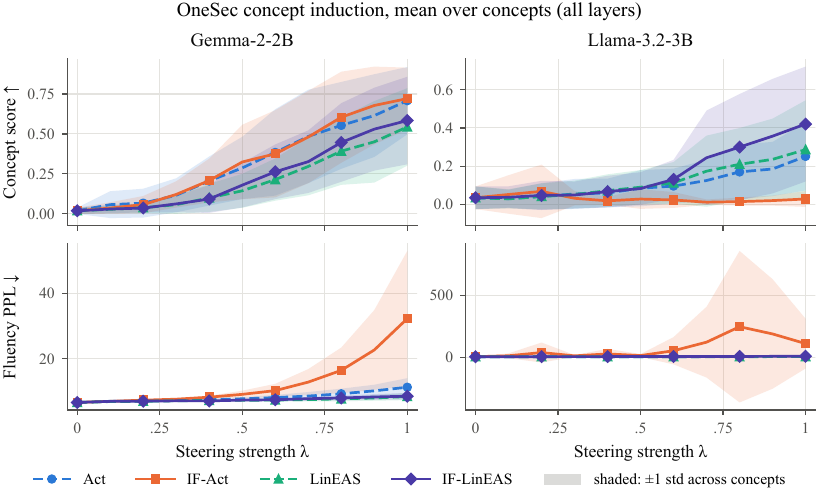}
  \caption{\textbf{IF weighted steering methods achieve the best concept indiction scores in most settings.} Apart from Llama IF-Act. We again see that while IF-Act gets good concept score performance it disrupts the model indiction much more than the other methods and causes increasingly poor perplexity scores. }
  \label{fig:fig_onesec_mean_all}
\end{figure}

\begin{figure}[t]
  \centering
\includegraphics[width=\linewidth]{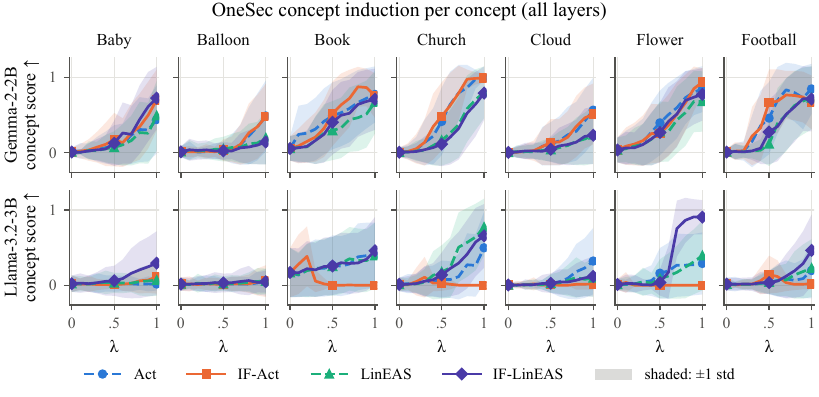}
  \caption{\textbf{Object based concepts achieve various levels of success.} While concepts such as 'church' , 'book', 'flower' and 'football' achieve good steering performance, the remaining concepts are steered relatively poorly, with Llama 'Balloon' completely failing any sort of concept indiction. }
  \label{fig:fig_onesec_grid_all}
\end{figure}

\begin{figure}[t]
  \centering
\includegraphics[width=\linewidth]{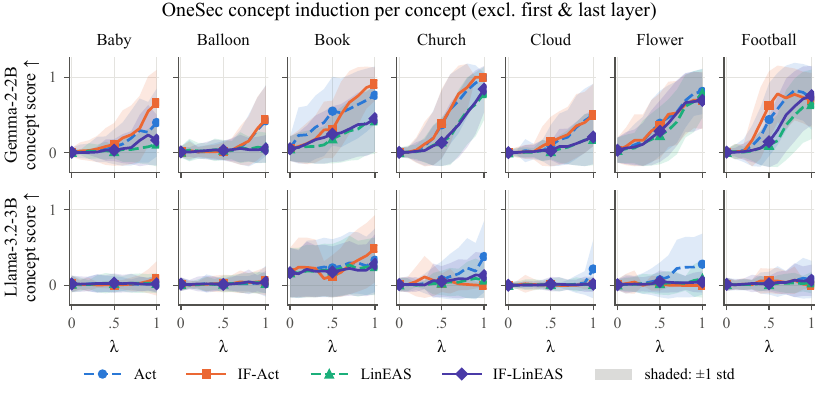}
  \caption{\textbf{Llama and Gemma's performance diverges when first and last layers are not steered upon} While Gemma adapts well to not steering on the first and last layers, Llama largely collapses and mostly fails to induce any concept meaningfully.}
  \label{fig:fig_onesec_grid_no_tb}
\end{figure}

\section{Full Toxicicty suppression Results}

\begin{table}[t]
  \caption{Toxicity suppression. Steered values at $\lambda$ = 1; the unsteered row is $\lambda$ = 0. Layers: \textit{All} = steering on all layers, \textit{Excl.} = first and last layer excluded. PPL columns: Flu. = fluency perplexity (of the generated continuation), Wiki = held-out WikiText perplexity (a side-effect check on general language modelling). RoB. = RoBERTa toxicity classifier judge. IF-LinEAS = weighted-Wasserstein LinEAS Full results with standard deviations.}
  \label{tab:toxicity-suppression-full}
  \centering
  \small
  \setlength{\tabcolsep}{3pt}
  \renewcommand{\arraystretch}{0.85}
  \setlength{\aboverulesep}{0.2ex}
  \setlength{\belowrulesep}{0.2ex}
  \resizebox{\textwidth}{!}{%
  \begin{tabular}{llccccccccc}
    \toprule
     &  & \multicolumn{2}{c}{\textbf{Toxicity} $\downarrow$} & \multicolumn{2}{c}{\textbf{PPL} $\downarrow$} &  & \multicolumn{2}{c}{\textbf{RTP} $\downarrow$} & \multicolumn{2}{c}{\textbf{TET} $\downarrow$} \\
    \cmidrule(lr){3-4} \cmidrule(lr){5-6} \cmidrule(lr){8-9} \cmidrule(lr){10-11}
    \textbf{Layers} & \textbf{Method} & \textbf{LLM} & \textbf{RoB.} & \textbf{Flu.} & \textbf{Wiki} & \textbf{MMLU} $\uparrow$ & \textbf{LLM} & \textbf{RoB.} & \textbf{LLM} & \textbf{RoB.} \\
    \midrule
    \multicolumn{11}{l}{\textit{Gemma-2-2B}} \\
    -- & Unsteered & 0.344 {\scriptsize$\pm$0.439} & 0.233 {\scriptsize$\pm$0.407} & 27.11 {\scriptsize$\pm$20.63} & 32.30 {\scriptsize(1.25)} & 0.550 {\scriptsize$\pm$0.498} & 0.107 {\scriptsize$\pm$0.309} & 0.033 {\scriptsize$\pm$0.179} & 0.219 {\scriptsize$\pm$0.413} & 0.133 {\scriptsize$\pm$0.340} \\
    \specialrule{0.25pt}{2pt}{2pt}
    \multicolumn{11}{l}{\textit{All layers}} \\
    All & Act & 0.083 {\scriptsize$\pm$0.252} & 0.044 {\scriptsize$\pm$0.187} & \textbf{27.45} {\scriptsize$\pm$20.41} & \textbf{32.43} {\scriptsize(1.23)} & \textbf{0.524} {\scriptsize$\pm$0.499} & \textbf{0.016} {\scriptsize$\pm$0.125} & \underline{0.005} {\scriptsize$\pm$0.071} & 0.098 {\scriptsize$\pm$0.298} & 0.048 {\scriptsize$\pm$0.214} \\
    All & IF-Act & \underline{0.031} {\scriptsize$\pm$0.145} & \textbf{0.001} {\scriptsize$\pm$0.019} & 31.13 {\scriptsize$\pm$20.09} & 51.60 {\scriptsize(1.34)} & 0.438 {\scriptsize$\pm$0.496} & \textbf{0.016} {\scriptsize$\pm$0.125} & \textbf{0.003} {\scriptsize$\pm$0.055} & \textbf{0.059} {\scriptsize$\pm$0.235} & \textbf{0.018} {\scriptsize$\pm$0.133} \\
    All & LinEAS & 0.095 {\scriptsize$\pm$0.271} & 0.045 {\scriptsize$\pm$0.195} & 27.69 {\scriptsize$\pm$20.16} & \underline{32.44} {\scriptsize(1.25)} & \underline{0.520} {\scriptsize$\pm$0.500} & 0.041 {\scriptsize$\pm$0.198} & 0.006 {\scriptsize$\pm$0.077} & 0.083 {\scriptsize$\pm$0.276} & 0.036 {\scriptsize$\pm$0.186} \\
    All & IF-LinEAS & \textbf{0.018} {\scriptsize$\pm$0.117} & \underline{0.009} {\scriptsize$\pm$0.086} & \underline{27.58} {\scriptsize$\pm$20.30} & 34.57 {\scriptsize(1.27)} & 0.520 {\scriptsize$\pm$0.500} & \underline{0.034} {\scriptsize$\pm$0.181} & 0.008 {\scriptsize$\pm$0.089} & \underline{0.073} {\scriptsize$\pm$0.260} & \underline{0.033} {\scriptsize$\pm$0.180} \\
    \midrule
    \multicolumn{11}{l}{\textit{Excl. first \& last layer}} \\
    Excl. & Act & 0.075 {\scriptsize$\pm$0.251} & 0.065 {\scriptsize$\pm$0.230} & \textbf{27.25} {\scriptsize$\pm$20.48} & 31.83 {\scriptsize(1.23)} & 0.531 {\scriptsize$\pm$0.499} & \underline{0.024} {\scriptsize$\pm$0.153} & \textbf{0.002} {\scriptsize$\pm$0.045} & 0.115 {\scriptsize$\pm$0.319} & 0.056 {\scriptsize$\pm$0.230} \\
    Excl. & IF-Act & \textbf{0.023} {\scriptsize$\pm$0.132} & \textbf{0.008} {\scriptsize$\pm$0.078} & 27.42 {\scriptsize$\pm$20.39} & 35.29 {\scriptsize(1.27)} & 0.466 {\scriptsize$\pm$0.499} & \textbf{0.015} {\scriptsize$\pm$0.122} & \underline{0.003} {\scriptsize$\pm$0.055} & \textbf{0.077} {\scriptsize$\pm$0.267} & \textbf{0.029} {\scriptsize$\pm$0.169} \\
    Excl. & LinEAS & 0.071 {\scriptsize$\pm$0.234} & 0.035 {\scriptsize$\pm$0.169} & 27.60 {\scriptsize$\pm$20.23} & \underline{30.96} {\scriptsize(1.22)} & \textbf{0.541} {\scriptsize$\pm$0.498} & 0.038 {\scriptsize$\pm$0.191} & 0.010 {\scriptsize$\pm$0.099} & 0.098 {\scriptsize$\pm$0.297} & 0.059 {\scriptsize$\pm$0.235} \\
    Excl. & IF-LinEAS & \underline{0.029} {\scriptsize$\pm$0.144} & \underline{0.010} {\scriptsize$\pm$0.087} & \underline{27.27} {\scriptsize$\pm$20.43} & \textbf{30.65} {\scriptsize(1.22)} & \underline{0.536} {\scriptsize$\pm$0.499} & 0.037 {\scriptsize$\pm$0.189} & 0.008 {\scriptsize$\pm$0.089} & \underline{0.082} {\scriptsize$\pm$0.275} & \underline{0.053} {\scriptsize$\pm$0.224} \\
    \midrule
    \multicolumn{11}{l}{\textit{Llama-3.2-3B}} \\
    -- & Unsteered & 0.268 {\scriptsize$\pm$0.416} & 0.219 {\scriptsize$\pm$0.400} & 25.06 {\scriptsize$\pm$21.27} & 24.20 {\scriptsize(0.94)} & 0.562 {\scriptsize$\pm$0.496} & 0.123 {\scriptsize$\pm$0.328} & 0.036 {\scriptsize$\pm$0.186} & 0.224 {\scriptsize$\pm$0.417} & 0.232 {\scriptsize$\pm$0.422} \\
    \specialrule{0.25pt}{2pt}{2pt}
    \multicolumn{11}{l}{\textit{All layers}} \\
    All & Act & 0.257 {\scriptsize$\pm$0.399} & 0.100 {\scriptsize$\pm$0.290} & \textbf{24.57} {\scriptsize$\pm$21.60} & 26.07 {\scriptsize(0.96)} & 0.517 {\scriptsize$\pm$0.500} & 0.071 {\scriptsize$\pm$0.257} & 0.019 {\scriptsize$\pm$0.137} & 0.193 {\scriptsize$\pm$0.395} & 0.166 {\scriptsize$\pm$0.372} \\
    All & IF-Act & 0.177 {\scriptsize$\pm$0.346} & 0.073 {\scriptsize$\pm$0.249} & 27.79 {\scriptsize$\pm$21.86} & 34.74 {\scriptsize(0.96)} & 0.276 {\scriptsize$\pm$0.447} & 0.064 {\scriptsize$\pm$0.245} & 0.014 {\scriptsize$\pm$0.117} & 0.195 {\scriptsize$\pm$0.396} & 0.159 {\scriptsize$\pm$0.365} \\
    All & LinEAS & \underline{0.101} {\scriptsize$\pm$0.280} & \underline{0.031} {\scriptsize$\pm$0.160} & \underline{24.67} {\scriptsize$\pm$21.40} & \underline{25.51} {\scriptsize(0.95)} & \textbf{0.554} {\scriptsize$\pm$0.497} & \underline{0.047} {\scriptsize$\pm$0.212} & \underline{0.012} {\scriptsize$\pm$0.109} & \underline{0.115} {\scriptsize$\pm$0.319} & \underline{0.120} {\scriptsize$\pm$0.324} \\
    All & IF-LinEAS & \textbf{0.052} {\scriptsize$\pm$0.202} & \textbf{0.017} {\scriptsize$\pm$0.110} & 25.76 {\scriptsize$\pm$20.83} & \textbf{25.06} {\scriptsize(0.95)} & \underline{0.554} {\scriptsize$\pm$0.497} & \textbf{0.039} {\scriptsize$\pm$0.194} & \textbf{0.007} {\scriptsize$\pm$0.083} & \textbf{0.097} {\scriptsize$\pm$0.296} & \textbf{0.112} {\scriptsize$\pm$0.316} \\
    \midrule
    \multicolumn{11}{l}{\textit{Excl. first \& last layer}} \\
    Excl. & Act & 0.214 {\scriptsize$\pm$0.380} & 0.143 {\scriptsize$\pm$0.342} & 24.65 {\scriptsize$\pm$21.54} & 24.93 {\scriptsize(0.94)} & 0.519 {\scriptsize$\pm$0.500} & 0.090 {\scriptsize$\pm$0.286} & 0.031 {\scriptsize$\pm$0.173} & 0.188 {\scriptsize$\pm$0.391} & 0.191 {\scriptsize$\pm$0.393} \\
    Excl. & IF-Act & 0.220 {\scriptsize$\pm$0.388} & 0.155 {\scriptsize$\pm$0.352} & \textbf{24.38} {\scriptsize$\pm$21.43} & 25.97 {\scriptsize(0.95)} & 0.524 {\scriptsize$\pm$0.499} & 0.094 {\scriptsize$\pm$0.292} & 0.031 {\scriptsize$\pm$0.173} & 0.205 {\scriptsize$\pm$0.404} & 0.196 {\scriptsize$\pm$0.397} \\
    Excl. & LinEAS & \underline{0.150} {\scriptsize$\pm$0.332} & \underline{0.076} {\scriptsize$\pm$0.250} & 24.89 {\scriptsize$\pm$21.21} & \underline{24.47} {\scriptsize(0.94)} & \underline{0.545} {\scriptsize$\pm$0.498} & \underline{0.049} {\scriptsize$\pm$0.216} & \underline{0.015} {\scriptsize$\pm$0.122} & \underline{0.133} {\scriptsize$\pm$0.340} & \underline{0.135} {\scriptsize$\pm$0.342} \\
    Excl. & IF-LinEAS & \textbf{0.061} {\scriptsize$\pm$0.208} & \textbf{0.027} {\scriptsize$\pm$0.148} & \underline{24.64} {\scriptsize$\pm$21.37} & \textbf{24.28} {\scriptsize(0.95)} & \textbf{0.553} {\scriptsize$\pm$0.497} & \textbf{0.045} {\scriptsize$\pm$0.207} & \textbf{0.009} {\scriptsize$\pm$0.094} & \textbf{0.118} {\scriptsize$\pm$0.322} & \textbf{0.120} {\scriptsize$\pm$0.325} \\
    \bottomrule
  \end{tabular}
  }
\end{table}

\begin{figure}[t]
  \centering
\includegraphics[width=\linewidth]{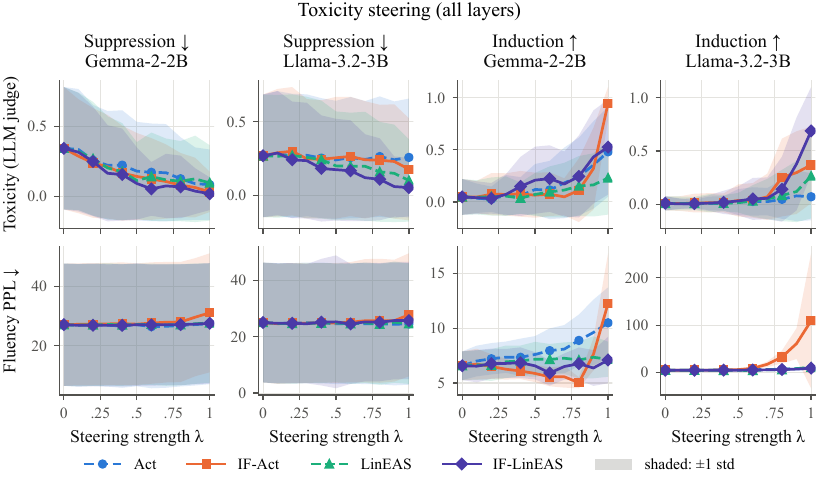}
  \caption{\textbf{IF weighted steering methods achieve the best concept suppression and induction scores, when compared to their non-weighted baselines.} Again we also see that while IF-Act achieves a strong toxicity indiction score, it does so at the cost of a very high perplexity score in both models. overall the suppression arms maintain a largely consistent perplexity, whilst this is largely explained by the lover overall change in its objective of suppression than the more volatile indiction objective.}
  \label{fig:fig_toxicity_all}
\end{figure}

\begin{figure}[t]
  \centering
\includegraphics[width=\linewidth]{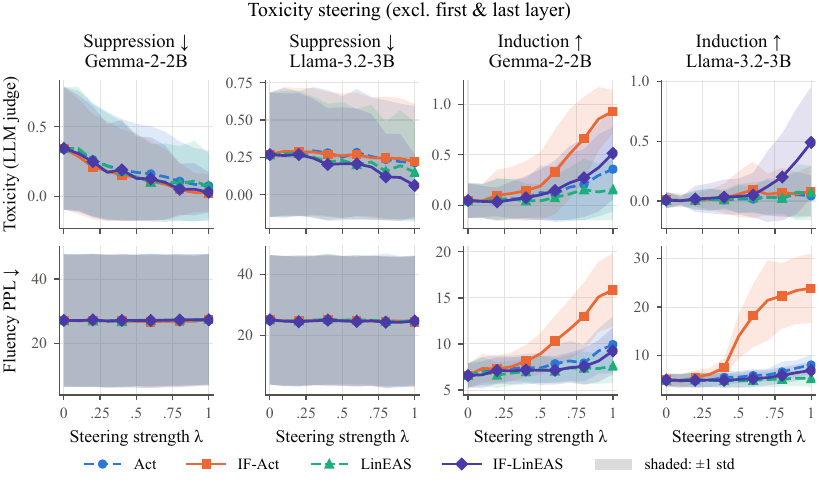}
  \caption{\textbf{As in Figure~\ref{fig:fig_toxicity_all}, IF-weighted steering methods
achieve the best concept suppression and induction scores in most settings,
with the exception of IF-Act on Llama supression.} IF-Act again reaches a strong toxicity induction
score, but only at the cost of very high perplexity in both models. The
perplexity trends also match the earlier results. Under the suppression
objective, perplexity stays roughly constant as steering strength increases.
Under the induction objective, it rises steadily.}
  \label{fig:fig_toxicity_no_tb}
\end{figure}

\section{Full Truthfulness induction Results}

\begin{table}[t]
  \caption{TruthfulQA: effect of the number of contrastive pairs. Steered values at $\lambda$ = 1; the unsteered row is $\lambda$ = 0. Layers: \textit{All} = steering on all layers, \textit{Excl.} = first and last layer excluded. PPL columns: Flu. = fluency perplexity (of the generated continuation), Wiki = held-out WikiText perplexity (a side-effect check on general language modelling). Fold-averaged MC1/MC2 at $\lambda$ = 1. $\Delta$ = n=1024 minus n=32. Full results, including layer-exclusion ablation.}
  \label{tab:truthfulqa-scaling-full}
  \centering
  \small
  \setlength{\tabcolsep}{4pt}
  \renewcommand{\arraystretch}{0.9}
  \setlength{\aboverulesep}{0.2ex}
  \setlength{\belowrulesep}{0.2ex}
  \begin{tabular}{llccccc}
    \toprule
     &  & \multicolumn{2}{c}{\textbf{MC1} $\uparrow$} & \multicolumn{2}{c}{\textbf{MC2} $\uparrow$} &  \\
    \cmidrule(lr){3-4} \cmidrule(lr){5-6}
    \textbf{Layers} & \textbf{Method} & \textbf{n=32} & \textbf{n=1024} & \textbf{n=32} & \textbf{n=1024} & \textbf{$\Delta$ MC2} $\uparrow$ \\
    \midrule
    \multicolumn{2}{l}{\textit{Gemma-2-2B (unsteered)}} & 0.209 {\scriptsize$\pm$0.407} & 0.209 {\scriptsize$\pm$0.407} & 0.336 {\scriptsize$\pm$0.395} & 0.336 {\scriptsize$\pm$0.395} & 0.000 \\
    \cmidrule(lr){1-7}
    \multirow{4}{*}{All} & Act & \underline{0.245} {\scriptsize$\pm$0.429} & \textbf{0.256} {\scriptsize$\pm$0.436} & \underline{0.396} {\scriptsize$\pm$0.413} & \underline{0.400} {\scriptsize$\pm$0.406} & 0.003 \\
     & IF-Act & \textbf{0.278} {\scriptsize$\pm$0.448} & \underline{0.255} {\scriptsize$\pm$0.436} & \textbf{0.459} {\scriptsize$\pm$0.449} & \textbf{0.482} {\scriptsize$\pm$0.455} & \textbf{0.023} \\
     & LinEAS & 0.212 {\scriptsize$\pm$0.408} & 0.214 {\scriptsize$\pm$0.410} & 0.348 {\scriptsize$\pm$0.396} & 0.353 {\scriptsize$\pm$0.398} & \underline{0.005} \\
     & IF-LinEAS & 0.229 {\scriptsize$\pm$0.420} & 0.203 {\scriptsize$\pm$0.402} & 0.370 {\scriptsize$\pm$0.400} & 0.356 {\scriptsize$\pm$0.405} & -0.013 \\
    \cmidrule(lr){1-7}
    \multirow{4}{*}{Excl.} & Act & \underline{0.236} {\scriptsize$\pm$0.424} & \textbf{0.256} {\scriptsize$\pm$0.436} & \underline{0.389} {\scriptsize$\pm$0.414} & \underline{0.393} {\scriptsize$\pm$0.406} & \underline{0.004} \\
     & IF-Act & \textbf{0.259} {\scriptsize$\pm$0.438} & \underline{0.244} {\scriptsize$\pm$0.429} & \textbf{0.452} {\scriptsize$\pm$0.449} & \textbf{0.471} {\scriptsize$\pm$0.451} & \textbf{0.020} \\
     & LinEAS & 0.211 {\scriptsize$\pm$0.408} & 0.218 {\scriptsize$\pm$0.413} & 0.346 {\scriptsize$\pm$0.395} & 0.350 {\scriptsize$\pm$0.397} & 0.004 \\
     & IF-LinEAS & 0.222 {\scriptsize$\pm$0.415} & 0.208 {\scriptsize$\pm$0.406} & 0.366 {\scriptsize$\pm$0.400} & 0.347 {\scriptsize$\pm$0.400} & -0.019 \\
    \midrule
    \multicolumn{2}{l}{\textit{Llama-3.2-3B (unsteered)}} & 0.245 {\scriptsize$\pm$0.430} & 0.245 {\scriptsize$\pm$0.430} & 0.396 {\scriptsize$\pm$0.414} & 0.396 {\scriptsize$\pm$0.414} & 0.000 \\
    \cmidrule(lr){1-7}
    \multirow{4}{*}{All} & Act & \underline{0.274} {\scriptsize$\pm$0.446} & \textbf{0.284} {\scriptsize$\pm$0.451} & \underline{0.448} {\scriptsize$\pm$0.420} & \underline{0.447} {\scriptsize$\pm$0.417} & -0.001 \\
     & IF-Act & \textbf{0.301} {\scriptsize$\pm$0.459} & \underline{0.264} {\scriptsize$\pm$0.441} & \textbf{0.468} {\scriptsize$\pm$0.444} & \textbf{0.474} {\scriptsize$\pm$0.447} & \textbf{0.006} \\
     & LinEAS & 0.256 {\scriptsize$\pm$0.436} & 0.256 {\scriptsize$\pm$0.436} & 0.401 {\scriptsize$\pm$0.414} & 0.402 {\scriptsize$\pm$0.415} & \underline{0.002} \\
     & IF-LinEAS & 0.257 {\scriptsize$\pm$0.437} & 0.253 {\scriptsize$\pm$0.435} & 0.408 {\scriptsize$\pm$0.416} & 0.401 {\scriptsize$\pm$0.415} & -0.007 \\
    \cmidrule(lr){1-7}
    \multirow{4}{*}{Excl.} & Act & \underline{0.269} {\scriptsize$\pm$0.444} & \textbf{0.280} {\scriptsize$\pm$0.449} & \underline{0.442} {\scriptsize$\pm$0.420} & \underline{0.447} {\scriptsize$\pm$0.417} & \textbf{0.005} \\
     & IF-Act & \textbf{0.301} {\scriptsize$\pm$0.459} & \underline{0.267} {\scriptsize$\pm$0.442} & \textbf{0.472} {\scriptsize$\pm$0.446} & \textbf{0.473} {\scriptsize$\pm$0.446} & \underline{0.001} \\
     & LinEAS & 0.258 {\scriptsize$\pm$0.437} & 0.253 {\scriptsize$\pm$0.435} & 0.402 {\scriptsize$\pm$0.415} & 0.402 {\scriptsize$\pm$0.415} & 0.000 \\
     & IF-LinEAS & 0.258 {\scriptsize$\pm$0.437} & 0.253 {\scriptsize$\pm$0.435} & 0.410 {\scriptsize$\pm$0.416} & 0.402 {\scriptsize$\pm$0.415} & -0.008 \\
    \bottomrule
  \end{tabular}
\end{table}

\clearpage

\clearpage

\section{Influence Score distributions}
\label{sec:Influence Score distributions}

\begin{figure}[t]
  \centering
\includegraphics[width=\linewidth]{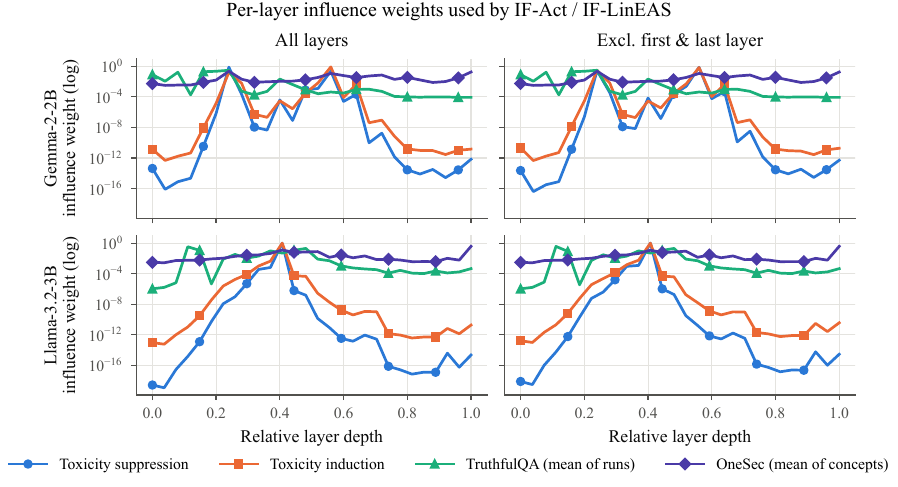}
  \caption{\textbf{Influence functions localise the toxicity concept to the middle layers.}
We plot the mean influence weight that each data point assigns to each layer of
the model, for each concept. A non-uniform distribution means that the influence
score has identified layers where steering on that data point is more beneficial.
For OneSec (object-based concepts) and TruthfulQA (truthfulness), the weights
are close to uniform across layers. For toxicity, they are clearly concentrated
in the middle layers.}
  \label{fig:fig_influence_layers}
\end{figure}
\clearpage

\clearpage

\section{Influence - Projection Proxy}
\label{sec:Influence-Projection Proxy}

\begin{table}[t]
  \caption{Influence vs. projection proxy: method-pair agreement (Gemma-2-2B, toxic). $\rho$\textit{ is the Spearman correlation between two methods' scores, corrected for both methods' split-half unreliability.}}
  \label{tab:influence-vs-projection-gemma-2-2b-toxic}
  \centering
  \small
  \setlength{\tabcolsep}{4pt}
  \renewcommand{\arraystretch}{0.9}
  \setlength{\aboverulesep}{0.2ex}
  \setlength{\belowrulesep}{0.2ex}
  \begin{tabular}{lcc}
    \toprule
    \textbf{Method pair} & \textbf{$\rho$* (candidate axis)} & \textbf{$\rho$* (layer axis)} \\
    \midrule
    \multicolumn{3}{l}{\textit{Core methods}} \\
    Projection proxy vs. Gradient norm & 0.039 & 0.172 \\
    EK-FAC influence vs. Projection proxy & 0.317 & 0.504 \\
    EK-FAC influence vs. Gradient norm & 0.079 & 0.346 \\
    \bottomrule
  \end{tabular}
\end{table}

\clearpage

\end{document}